\documentclass[10pt,journal,compsoc]{IEEEtran}
\usepackage{graphicx} % DO NOT CHANGE THIS

\usepackage{amsmath}
\usepackage{amssymb}
\usepackage{booktabs}
\usepackage{multirow}
\usepackage{xcolor}
\usepackage{bbm}
\usepackage{array}
\usepackage{ragged2e}

\definecolor{citecolor}{HTML}{0071bc}
\usepackage[pagebackref=false,breaklinks=true,colorlinks,citecolor=citecolor,bookmarks=false]{hyperref}

\newlength\savewidth
\newcommand{\tablestyle}[2]{\setlength{\tabcolsep}{#1}\renewcommand{\arraystretch}{#2}\centering\footnotesize}
\makeatletter\renewcommand\paragraph{\@startsection{paragraph}{4}{\z@}
  {.5em \@plus1ex \@minus.2ex}{-.5em}{\normalfont\normalsize\bfseries}}\makeatother

\ifCLASSOPTIONcompsoc
  \usepackage[nocompress]{cite}
\else
  \usepackage{cite}
\fi

\newcommand{\ie}{\textit{i}.\textit{e}.}
\newcommand{\eg}{\textit{e}.\textit{g}.}
\newcommand{\etal}{\textit{et al}. }

\begin{document}

\title{End-to-end One-shot Human Parsing}

\author{Haoyu~He, 
Jing~Zhang,~\IEEEmembership{Senior Member,~IEEE,}
Bohan~Zhuang,
Jianfei~Cai,~\IEEEmembership{Fellow,~IEEE,}
and Dacheng~Tao,~\IEEEmembership{Fellow,~IEEE}
% <-this % stops a space
% All authors have extensively deliberated on each individual's contributions throughout the entire project and have reached a unanimous consensus on the finalized list of authors. Kindly incorporate this approved author list in the final version of the work.
\IEEEcompsocitemizethanks{
\IEEEcompsocthanksitem This work was supported by ARC FL170100117. This work was partly done when H.~He was a Master of Philosophy student at The University of Sydney. Corresponding authors: Jing Zhang; Bohan Zhuang.
\protect\\
\IEEEcompsocthanksitem H.~He, B.~Zhuang and J.~Cai are with the Department of Data Science and AI, Faculty of IT, Monash University, Australia.
E-mail: Haoyu.He@monash.edu, Bohan.Zhuang@monash.edu, and \mbox{Jianfei.Cai@monash.edu}
\IEEEcompsocthanksitem J.~Zhang and D.~Tao are with the Faculty of Engineering, The University of Sydney, Australia. E-mail: Jing.Zhang1@sydney.edu.au and Dacheng.Tao@gmail.com
}% <-this % stops an unwanted space
}

\markboth{}%
{Shell \MakeLowercase{\textit{et al.}}: Bare Demo of IEEEtran.cls for Computer Society Journals}

\IEEEtitleabstractindextext{%
\begin{abstract}
\justifying
Previous human parsing models are limited to parsing humans into pre-defined classes, which is inflexible for practical fashion applications that often have new fashion item classes. In this paper, we define a novel one-shot human parsing (OSHP) task that requires parsing humans into an open set of classes defined by any test example. During training, only base classes are exposed, which only overlap with part of the test-time classes. To address three main challenges in OSHP, \ie, small sizes, testing bias, and similar parts, we devise an End-to-end One-shot human Parsing Network (EOP-Net). Firstly, an end-to-end human parsing framework is proposed to parse the query image into both coarse-grained and fine-grained human classes, which builds a strong embedding network with rich semantic information shared across different granularities, facilitating identifying small-sized human classes. Then, we propose learning momentum-updated prototypes by gradually smoothing the training time static prototypes, which helps stabilize the training and learn robust features. Moreover, we devise a dual metric learning scheme which encourages the network to enhance features' representational capability in the early training phase and improve features' transferability in the late training phase. Therefore, our EOP-Net can learn representative features that can quickly adapt to the novel classes and mitigate the testing bias issue. In addition, we further employ a contrastive loss at the prototype level, thereby enforcing the distances among the classes in the fine-grained metric space and discriminating the similar parts. To comprehensively evaluate the OSHP models, we tailor three existing popular human parsing benchmarks to the OSHP task. Experiments on the new benchmarks demonstrate that EOP-Net outperforms representative one-shot segmentation models by large margins, which serves as a strong baseline for further research on this new task. The source code is available at \url{https://github.com/Charleshhy/One-shot-Human-Parsing}.

\end{abstract}

}

\maketitle

\IEEEdisplaynontitleabstractindextext
\IEEEpeerreviewmaketitle

\IEEEraisesectionheading{\section{Introduction}\label{sec:introduction}}

\IEEEPARstart{H}{uman} parsing is a fundamental visual understanding task, requiring segmenting human instances into explicit body parts as well as some clothing classes at the pixel level. It has a broad range of downstream applications, such as fashion image generation~\cite{han2019clothflow}, virtual try-on~\cite{dong2019fw,wu2019m2e}, and fashion image retrieval~\cite{wang2017clothing}. Recent efforts in Convolutional Neural Networks (CNN) based solutions have achieved significant progress by leveraging large-scale human parsing datasets with fine-grained human class annotations. However, the parsing capability is accordingly restricted to the classes pre-defined in the training set, \eg, 18 classes in ATR~\cite{liang2015}, 20 classes in CIHP~\cite{gong2018instance}, and 20 classes in LIP~\cite{liang2018look}. Due to the vast new clothing, fast varying styles, and various trending outfits in the fashion industry, parsing humans into fixed and pre-defined classes has limited the usage of human parsing models in the wide range of downstream applications.

\begin{figure*}[t]
  \centering
  \includegraphics[width=\linewidth]{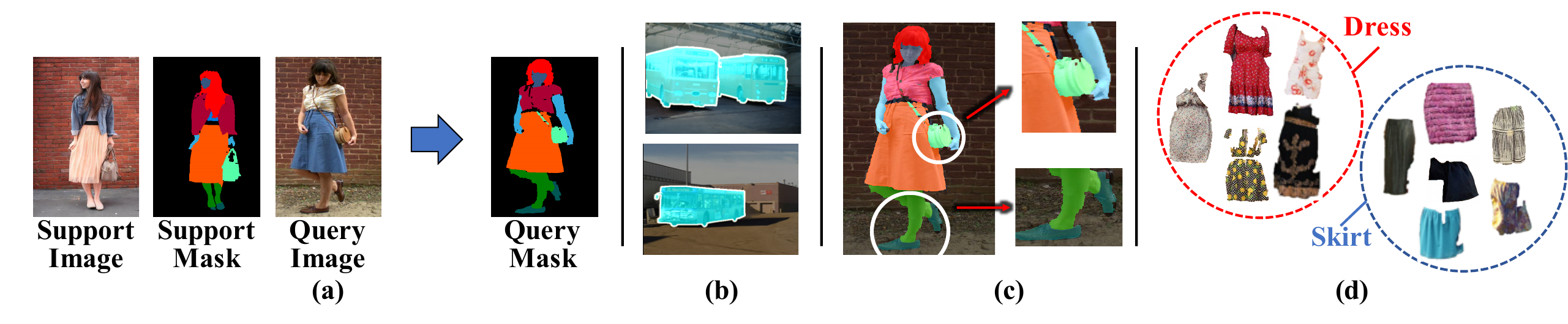}
  \vspace{-2.5em}
  \caption{(a) Illustration of the proposed One-Shot Human Parsing
  (OSHP) task. OSHP requires parsing the target query image into an open set of classes defined in the support mask. (b) The classes in One-Shot Semantic Segmentation
  (OS3) are large and holistic objects. (c) The classes in OSHP are small and entangled with others.
  (d) In OSHP, the fine-grained human classes are similar and require higher discriminative capability. Best viewed in color. 
  }\vspace{-1.5em}
  \label{fig:intro}
\end{figure*}

Inspired by the progress in one-shot learning~\cite{koch2015siamese, vinyals2016matching}, we make the first attempt to solve the aforementioned problem by defining a new task named One-Shot Human Parsing (OSHP), illustrated in Figure~\ref{fig:intro} (a). OSHP requires parsing humans in a query image into an open set of reference classes. The classes are defined by any single reference example (\ie, a support image) during testing, no matter whether they are annotated during training (denoted as base classes) or not (denoted as novel classes). In this way, the novel classes with different semantics to the base classes can be flexibly added, removed, and re-organized depending on specific application requirements. 
Accordingly, there is no need for collecting and annotating new training samples and retraining the parsing models. 

One similar task is One-Shot Semantic Segmentation (OS3)~\cite{zhang2019canet, zhang2019pyramid, wang2019panet}, which requires transferring the segmentation knowledge from the pre-defined base classes to the novel classes as shown in Figure~\ref{fig:intro} (b). However, OSHP is different from OS3 in three ways, which accordingly
delivers
three key challenges. Firstly, the objects in OS3 to be segmented are mostly intact and salient foreground. In contrast, the human classes that need to be recognized in OSHP are small and entangled with the other parts, \ie, the issue of small sizes. The differences are shown in Figure~\ref{fig:intro} (c). Secondly, during testing, OS3 only evaluates the novel classes where the base classes are rarely shown and do not bias the evaluation. 
However,
OSHP requires recognizing both base classes and novel classes simultaneously during testing, which is a variant of generalized few-shot learning (GFSL) problem~\cite{gidaris2018dynamic, ren2019incremental, shi2019relational, ye2019learning, tian2020generalized}. Note that the two types of classes have highly imbalanced data, \ie, the model is trained with abundant base classes annotations but no novel classes annotations. Moreover, since we have no prior information nor an explicit definition of the novel classes during training, they are naturally annotated as background in the training data. Consequently, the parsing model may overfit the base classes and explicitly lean towards the background class for those novel classes, leading to the testing bias issue. Finally, in contrast to the unique objects in OS3, the human foreground consists of many similar parts, especially for the different fashion items, which are similar in shapes and relative positions within the human body, \eg, dress and skirt in Figure~\ref{fig:intro} (d). Consequently, when directly deploying the state-of-the-art OS3 models to OSHP, there will be a significant performance drop attributed to the aforementioned issues.

In this work, we propose a novel End-to-end One-shot human Parsing Network (EOP-Net) for OSHP. Following the metric-learning scheme that is widely adopted in One/Few-shot Learning~\cite{snell2017prototypical, vinyals2016matching}, EOP-Net performs metric learning on two explicit metric spaces derived from the same backbone encoder. Specifically, one space discriminates the salient human foreground, and the other discriminates the fine-grained human classes. Therefore, we efficiently share the semantic information across different granularities and build a strong embedding network, which helps segment the small-sized human classes. Then, we propose to gradually smooth the training time prototypes to obtain more stable and discriminative momentum-updated prototypes as the human class representations. The momentum-updated prototypes facilitate stabilizing training and learning robust features in the OSHP task. Besides, we propose to perform prototype learning~\cite{snell2017prototypical}
with a novel dual metric learning (DML) scheme. 
In DML, we design an Attention Guidance Module (AGM) that boosts features' representational capability in the early training phase and a Nearest Prototype Module (NPM) that enhances the model's transferability in the late training phase. The two modules are integrated seamlessly with a weight-shifting strategy to reduce the testing bias issue. 
In addition to supervising the predictions by pixels, we propose an auxiliary prototype-level contrastive loss so that the similar human classes are separated in the metric space.

To facilitate benchmarking our EOP-Net and other OSHP models for future studies, we specifically tailor three popular large-scale human parsing datasets ATR~\cite{liang2015}, LIP~\cite{gong2017look} and CIHP~\cite{gong2018instance} to 
the proposed OSHP task. There are two one-shot settings for each dataset: parsing one human class each time and parsing multiple human classes each time, \ie, $1$-way OSHP and $k$-way OSHP. Besides, the three tailored datasets cover a variety of scenes with humans in large appearance diversity, extreme poses, and occlusion. These datasets also include both single-person and multi-person scenarios. We believe that these three tailored datasets can provide comprehensive evaluations and train better OSHP models which can generalize to a wide range of one-shot applications.

The main contributions of this work are as follows:

\begin{itemize}
\vspace{-0.5em}
    \item We define a new and challenging task, \ie, One-Shot Human Parsing, which brings new challenges and insights to the human parsing and one-shot learning communities.
    
    \item To address the OSHP problem, we propose a novel one-shot human parsing method named EOP-Net that is built upon an end-to-end human parsing framework to learn a strong embedding network, momentum-updated prototypes that stabilize training, a DML scheme that simultaneously boosts features' representational capability and transferability, and a prototype-level contrastive loss to separate similar classes. The proposed method can efficiently address the three key challenges in OSHP, \ie, small sizes, testing bias, and similar parts, and produce high-quality predictions.
    
    \item We specifically tailor three large-scale human parsing datasets to suit the challenging OSHP task and facilitate benchmarking different OSHP models.
    
    \item Extensive experiments on the three datasets demonstrate that our EOP-Net achieves superior performance that outperforms the representative OS3 models by large margins and can serve as a strong baseline for the new OSHP task.
    
\end{itemize}

This work is built upon our earlier conference paper~\cite{he2021progressive}. In~\cite{he2021progressive}, we made the first attempt to formulate the challenging OSHP problem and proposed a POPNet with a progressive parsing framework and a dual metric learning scheme to address two main challenges in OSHP: small sizes and testing bias. In this work, we extend the preliminary version from several aspects. 1) We devise a novel OSHP model named EOP-Net, which adopts an end-to-end human parsing framework and reduces the POPNet's computational complexity and network parameters by half. EOP-Net also significantly improves the parsing performance for both base and novel classes from POPNet. 2) We identify the similar parts issue that significantly degrades the POPNet's discriminative ability. Therefore, we employ a prototype-level contrastive loss to remedy the issue and improve the model's discriminative capability. 3) We tailor two more human parsing datasets with more complicated human scenes into our OSHP setting to construct a comprehensive benchmark. 4) We apply EOP-Net, POPNet, and other SOTA OS3 methods implemented by us to these datasets to construct strong baselines for the new task. 5) We observe that building the momentum-updated prototypes for the base classes can improve features' representational capability, which is transferable to the novel concepts and can further boost the parsing performance. 6) We conduct more ablative studies and analyze more qualitative results in different dimensions to further investigate the effectiveness of our methods.
\vspace{-1em}
\section{Related Work}
\subsection{Human Parsing}
Human parsing aims at segmenting an image containing humans into semantic sub-parts, including body classes and clothing classes at the pixel level. Many efforts have been made in human parsing because of its wide range of downstream applications, especially in the fashion industry.

Recent advances in deep neural networks have made great progress in the semantic segmentation task~\cite{ronneberger2015u, chen2017deeplab} and the human parsing task~\cite{li2017multiple, zhao2017self, luo2018macro}. Since the human body contains highly structural information, many previous methods enhance the pixel-level representations with well-designed architectures that can capture the global context cues, such as global context embeddings~\cite{ruan2019devil, liang2015}, generative adversarial networks~\cite{luo2018macro, li2020multi}, 
and recurrent models \cite{liang2016semantic, liang2016semantic2}. 
Apart from pixel-level semantics, human classes naturally have rich structural semantics. Hence, many works model the human class correlations explicitly by building, \eg, graph neural networks~\cite{zhang2020correlating, he2020grapy}, tree-like topology message passing architectures~\cite{wang2019learning, ji2019learning}, and hierarchical human structures~\cite{zhu2018progressive, zhang2020part, li2020self}. Another direction is exploiting common semantics among different human-centric tasks, \eg, pose estimation and keypoint detection~\cite{fang2018weakly, nie2018mutual, xia2017joint, nie2018human, fang2018weakly, dong2014towards, liang2018look} or other prior human semantics, \eg, edge information or human contour~\cite{zhang2020correlating, li2020self2}. In addition to these methods that focus on modeling human semantics, many efforts specifically contribute to instance-aware settings~\cite{li2017multiple, li2018multi, zhao2020fine, yang2019parsing, yang2020renovating}.

Although achieving promising parsing results, the current methods are limited to parsing a fixed set of classes pre-defined in the training data, severely limiting the adaptation ability of human parsing models to scenarios requiring parsing new classes. Recently, \cite{gong2019graphonomy, he2020grapy} explore universal human parsing that is capable of parsing humans into multiple semantic label sets by training on multiple datasets simultaneously and transferring similar classes' semantics across different domains. However, it is still non-trivial to add new classes or re-organize existing classes without re-training or heavy fine-tuning. In contrast to the previous work, we make the first attempt to propose one-shot human parsing that requires parsing humans into an open set of classes including both the pre-defined base classes and the novel classes without collecting and annotating new training samples.

\vspace{-1em}
\subsection{Few-Shot Semantic Segmentation}\label{subsec:rel_os3}
One-Shot Semantic Segmentation (OS3) \cite{shaban2017one} aims to segment the novel objects from the query image by referring to a single support reference (support image and the support object mask). Following the one/few-shot learning \cite{koch2015siamese, finn2017model, snell2017prototypical, sung2018learning, chen2020new, liu2020crnet, tian2020prior}, a typical OS3 solution is to learn a good metric that can encode pixels from the same class close in the embedding space \cite{zhang2018sg, rakelly2018conditional, hu2019attention, tian2020differentiable, lu2021simpler, wu2021learning, yang2021mining}. For instance, SG-One~\cite{zhang2018sg} extracts the target class centroid and calculates the cosine similarity scores as attention to enhance the metric quality. MM-Net~\cite{wu2021learning} boosts the features with a set of meta-class memory to improve the model's generalization capability. To make full use of the correlations between the query and the support images, some other works refine the query and support features from their counterpart by cross-referencing~\cite{zhang2019canet}, graph convolution~\cite{zhang2019pyramid}, and graph attentions~\cite{cao2020few}. 

The methods mentioned above focus on segmenting one class from the query image at one time, while a more general segmentation setting is segmenting $k$ classes at the same time, \ie, one-shot $k$-way semantic segmentation~\cite{dong2018few, wang2019panet, siam2019amp}. For example, Dong \etal~\cite{dong2018few} propose to extract the class prototype and predict a probability map for each class, then fuse the prototypes into a complete $k$-class prediction. Liu \etal~\cite{liu2020part} further decompose the class representations into part-aware prototypes to capture fine-grained features. 

In contrast to the typical OS3 tasks where only novel classes are presented and required to be segmented, OSHP requires parsing humans into both base classes and novel classes simultaneously. OSHP is similar to the challenging generalized few-shot learning (GFSL) setting tailored for practical application scenarios \cite{gidaris2018dynamic, ren2019incremental, shi2019relational, ye2019learning}. 
In this paper, we make the first attempt to define the challenging OSHP task, construct a comprehensive benchmark tailored from the existing popular human parsing datasets, and propose a novel EOP-Net for OSHP. EOP-Net employs a DML scheme to enhance the transferability of the human parsing model for recognizing human classes from base classes to novel classes. Moreover, the human classes to be segmented in OSHP are small and similar, making the OSHP problem particularly challenging. To address these issues, we propose an end-to-end human parsing framework to locate the small classes and a prototype-level contrastive loss to separate the similar human classes.

\vspace{-1em}
\subsection{Contrastive Learning}
% \vspace{-0.5em}
The key idea of contrastive learning is pulling positive instances of the same semantic class closer and pushing away negative instances from the other classes. Tremendous efforts have been made in unsupervised representation learning~\cite{wu2018unsupervised, chen2020simple, ge2020self}, \eg, He \etal~\cite{he2020momentum} store the queue of negative samples in the memory as a dictionary look-up task and encode these samples with a momentum encoder. In fact, performing contrastive learning on instances with InfoNCE/Neighborhood Component Analysis (NCA) loss~\cite{gutmann2010noise, movshovitz2017no, goldberger2004neighbourhood} is comparable to using softmax classifiers with cosine or Euclidean distance according to~\cite{qi2018low}. To this end, SOTA OS3 methods attempt to minimize the contrastive InfoNCE/NCA loss on all the query pixels,
which may contain many noisy pixels, consequently hindering the model from learning a good metric. In contrast,
we propose a simple auxiliary contrastive loss on prototypes by grouping the pixel features 
from the same class. We empirically find that the contrastive loss on the grouped representations can equip the features with a higher discriminative capability, thereby improving the parsing performance.

\begin{table*}[t]
    \centering
    \footnotesize
    \caption{Notations describing the OSHP task and our EOP-Net.}\vspace{-1em}
    \resizebox{0.8\textwidth}{!}{%
    \begin{tabular}{l|c|l}
    \toprule
    Notation        & Type & Description  \\
    \midrule
    $|{\Lambda}^s_c|$  & scalar    & size of ${\Lambda}^s_c$ \\
    $|{\Lambda}^q_c|$  & scalar    & size of ${\Lambda}^q_c$ \\
    $\boldsymbol{p}_{{\rm cgs}, c}$ & vector & support static prototype for the $c$-th class in the coarse-grained metric space \\
    $\boldsymbol{p}_{{\rm fgs}, c}$ & vector & support static prototype for the $c$-th class in the fine-grained metric space \\
    $\boldsymbol{\tilde{p}}_{{\rm fgs},c}$ & vector & query static prototype for the $c$-th class \\
    $\boldsymbol{p}_{{\rm cgs}, c}^d$ & vector & support momentum-updated prototype for the $c$-th class in the coarse-grained metric space \\
     $\boldsymbol{p}_{{\rm fgs}, c}^d$ & vector & support momentum-updated prototype for the $c$-th class in the fine-grained metric space \\
    $ \boldsymbol{A}_c$ &  matrix   & cosine similarity map between the momentum-updated prototypes and residual features for the $c$-th class \\
    $ \boldsymbol{l}_c$ & matrix & AGM probability scores for the $c$-th class \\
    $\hat{\boldsymbol{M}}^{q;{\rm AGM}}_{c}$ &  matrix & predicted mask for the $c$-th class in the AGM module  \\
    $\hat{\boldsymbol{M}}^{q;{\rm NPM}}_{c}$ &  matrix & predicted mask for the $c$-th class in the NPM module  \\
    $\boldsymbol{I}^s$, $\boldsymbol{I}^{s'}$, $\boldsymbol{I}^q$, $\boldsymbol{I}^{q'}$  & tensor       & image \\
    $\boldsymbol{M}^s_{C_s}$, $\boldsymbol{M}^q_{C_s}$ &  tensor       & ground truth mask at fine-grained metric space for support and query, respectively \\
    $\hat{\boldsymbol{M}}^{q;{\rm AGM}}_{C_s}$ &  tensor       & predicted mask at fine-grained metric space in AGM \\
    $\hat{\boldsymbol{M}}^{q;{\rm NPM}}_{C_s}$ &  tensor       & predicted mask at fine-grained metric space in NPM \\
    $\boldsymbol{M}^s_{\rm bi}$, $\boldsymbol{M}^q_{\rm bi}$ &  tensor       & ground truth binary masks at coarse-grained metric space \\
    $ \boldsymbol{G}^s$ &  tensor   & support features encoded by the embedding network \\
    $ \boldsymbol{G}^q$ &  tensor   & query features encoded by the embedding network \\
    $ \boldsymbol{R}_c$ &  tensor   & refined features for the $c$-th class in the AGM \\
    $\mathcal{F}$   &  function  & mapping from the pair ((support image, support mask), query image) to query mask \\
    $g(;, \theta)$ &  function  & Siamese feature embedding network \\
    $f_{\rm cgs} (\cdot)$ &  function  & linear projection function for the coarse-grained metric space \\
    $f_{\rm fgs} (\cdot)$ &  function  & linear projection function for the fine-grained metric space \\
    $\varphi(\cdot)$, $\varphi_{bg}(\cdot)$ &  function  & several separable convolutional layers \\
    $\omega(\cdot)$, $\omega_{bg}(\cdot)$ &  function  & fully-connected projection layer \\
    $\mathcal{S}_{train}$   & set    & support set during meta-training \\
    $\mathcal{S}_{test}$   & set    & support set during meta-testing \\
    $\mathcal{Q}_{train}$   & set    & query set during meta-training \\
    $\mathcal{Q}_{test}$   & set    & query set during meta-testing \\
    $C_{novel}$     & set    & novel class set defined in dataset \\
    $C_{base}$     & set    & base class set defined in dataset \\
    $C_{human}$     & set    & the class set including all classes defined in dataset\\
    $C_{s}$, $C_{s'}$   & set    & the class set for the $s$ and $s'$-th support pair \\
    ${\Lambda}^s_c$  & set & the index set for pixels in the $c$-th class in the ground truth support mask \\
    ${\Lambda}^q_c$  & set  & the index set for pixels in the $c$-th class in the ground truth query mask \\
    \bottomrule
    \end{tabular}}\vspace{-1em}
    \label{tab:notations}
\end{table*}

\section{Problem Definition} \label{subsec:problemDef}

\noindent\textbf{Notations.} We denote scalars, vectors, matrices or tensors, and sets using lowercase, bold lowercase, bold uppercase, and uppercase (\eg, $\tau$, $\boldsymbol{p}$, $\boldsymbol{G}$, and $\mathcal{S}$), respectively. Please refer to Table~\ref{tab:notations}
for a vis-to-vis explanation of the notations we used.

In this paper, we propose a new task named OSHP that requires parsing humans into different semantic classes given a single dense annotated example \footnote{We do not expand to the few-shot human parsing scenario since a few support references make the human classes required to be parsed unbalanced in each episode, which overly complicates the training/evaluation process.}, and only part of the classes are labeled in the training data. We formulate OSHP as a meta-learning problem \cite{shaban2017one, vinyals2016matching, zhang2018sg} and train a meta-learner to solve randomly sampled OSHP episodes that require parsing different base class combinations. During testing, the meta-learner solves new episodes of parsing class combinations that include both the base classes and the novel classes. 

Formally, only the base class set $C_{base}$ is exposed in the meta-training phase. For one episode in meta-training, one support image-mask pair $(\boldsymbol{I}^s, \boldsymbol{M}^s_{C_s})$ and one query image $\boldsymbol{I}^q$ are randomly sampled from $\mathcal{S}_{train}$ and $\mathcal{Q}_{train}$, where $s$ and $q$ are indexes for the support set $\mathcal{S}_{train}$ and the query set $\mathcal{Q}_{train}$ during meta-training. Note that the class set $C_s\subseteq C_{base}$ contains the classes annotated in the $s$-th support pair and the query image is required to be parsed into the class set $C_s$. Here the meta-learner aims to learn a mapping $\mathcal{F}$ subjected to $\mathcal{F} \big((\boldsymbol{I}^s, \boldsymbol{M}^s_{C_s}), \boldsymbol{I}^q\big) \to \boldsymbol{M}^q_{C_s}$.

\begin{figure*}[t]
  \centering
  \includegraphics[width=\linewidth]{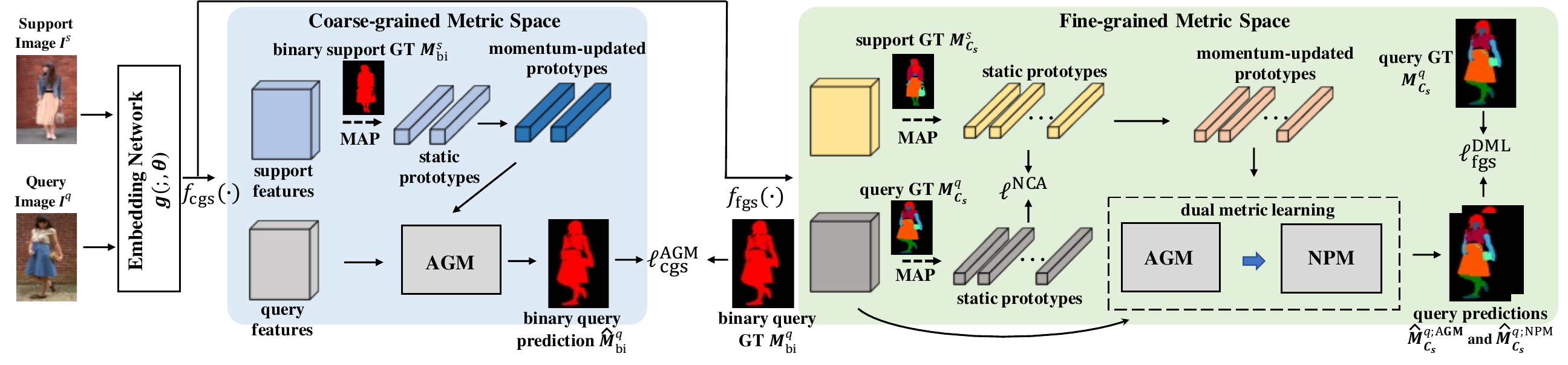}
  \vspace{-2em}
  \caption{The overview of the proposed EOP-Net, which contains a coarse-grained metric space (CGS) parsing images into binary human foreground masks and a fine-grained metric space (FGS) parsing images into the small human classes. Specifically, the embedded support and query features encoded by the embedding network are concurrently projected with $f_{\rm cgs}(\cdot)$ and $f_{\rm fgs}(\cdot)$. In CGS, we first generate the static prototypes with masked average pooling (MAP) as the class centroids for the human foreground and background. Then, we accumulate the momentum-updated prototypes and feed them into an Attention Guidance Module (AGM) to predict the human foreground from the query image. In FGS, we pool from both support and query features to get the static prototypes, which are regularized by a prototype-level contrastive loss $\ell^{\rm NCA}$. Next, we get the momentum-updated prototypes and perform prototype learning on the query features with a dual metric learning (DML) scheme, where we gradually shift the network's focus from an Attention Guidance Module to a Nearest Prototype Module (NPM) during training. Finally, we generate the fine-grained query prediction for both AGM and NPM that are supervised by a ground-truth query mask. See Section~\ref{subsec:end2end} for details.}\vspace{-1.5em}
  \label{fig:main}
\end{figure*}

While during meta-testing, both the base classes and the novel classes in the class set $C_{novel}$ are required to be parsed. For simplicity, we denote the target class set as $C_{human} = C_{base} \cup C_{novel}$. Similarly, for one episode in meta-testing, one support pair $(\boldsymbol{I}^{s'}, \boldsymbol{M}^{s'}_{C_{s'}})$ and one query image $\boldsymbol{I}^{q'}$ are randomly sampled from the test-time support set $\mathcal{S}_{test}$ and the query set $\mathcal{Q}_{test}$. Here $s'$ and $q'$ are respectively the indexes for $\mathcal{S}_{test}$ and $\mathcal{Q}_{test}$, and $C_{s'}$ is the class set for the $s'$-th support pair such that $C_{s'}\subseteq C_{human}$. In the meta-testing phase, the meta-learner quickly adapts to the new pairs, \ie,  $\mathcal{F} \big(({\boldsymbol{I}^{s'}}, {\boldsymbol{M}^{s'}_{C_{s'}}}), {\boldsymbol{I}^{q'}}\big) \to {\boldsymbol{M}^{q'}_{C_{s'}}}$.

\vspace{-1em}
\section{Method}
To address the challenging OSHP task, we devise an EOP-Net to embody the meta-learner via metric learning. The overview for our EOP-Net is depicted in Figure~\ref{fig:main}. Specifically, we first introduce our end-to-end human parsing framework (Section~\ref{subsec:end2end}), where we perform metric learning on a coarse-grained space discriminating the salient human foreground and a fine-grained space discriminating the human classes simultaneously. Metric learning on the coarse-grained metric space facilitates eliminating the non-important areas and focusing on the small-sized classes within the human foreground. Then, we propose to construct robust momentum-updated prototypes instead of the conventional static prototypes that are generated from a single support reference 
(Section~\ref{subsec:dp}). Next, for the coarse-grained metric space, we propose an AGM module (Section~\ref{subsec:agm}) to enhance the representational capability for the features. Considering AGM can easily overfit to only the base classes, we develop an NPM module (Section~\ref{subsec:npm}) that has higher transferability to the novel classes in the fine-grained metric space. We design a novel DML scheme that remedies the testing bias issue by integrating the AGM module with the NPM module via a weight-shifting strategy. Finally, a prototype-level contrastive loss (Section~\ref{subsec:contrastive}) is proposed to separate the similar parts in the metric space.

\subsection{End-to-End Human Parsing Framework}
\label{subsec:end2end}
Instead of being intact objects, human parsing classes are non-holistic small human parts, which challenges gathering the local semantic information and identifying the small classes for the human parsing models. Therefore, directly adapting the OS3 metric learning methods to the OSHP task degrades the performance significantly. Instead, we propose a highly efficient end-to-end human parsing framework, which discriminates the salient human foreground (in coarse-grained metric space) and the small human classes (in fine-grained metric space) concurrently.

To perform parsing in the coarse-grained metric space, we first generate additional binary masks in the coarse-grained metric space for both support and query images with a general ``human foreground'' label consisting of all human class segments and a ``human background'' label consisting of the other segments. We denote the generated binary masks as $\boldsymbol{M}^s_{\rm bi}$ and $\boldsymbol{M}^q_{\rm bi}$ corresponding to the support and the query images, respectively. In this case, $\boldsymbol{M}^q_{\rm bi}$ is also employed to provide supervision signals in the coarse-grained metric space. These masks can be easily derived offline from a universal human foreground parser. 

Next, we get the encoded query and support features $\boldsymbol{G}^{q}$ and $\boldsymbol{G}^{s}$ through a Siamese feature embedding network $g(;, \theta)$, \ie, $\boldsymbol{G}^{q}=g(\boldsymbol{I}^q, \theta)$ and $\boldsymbol{G}^{s}=g(\boldsymbol{I}^s, \theta)$, where $\boldsymbol{I}^s$ and $\boldsymbol{I}^q$ are the support and query images. We employ the Xception backbone~\cite{chollet2017xception} pretrained on the COCO dataset~\cite{lin2014microsoft} as our shared embedding network for both query and support images. Notably, instead of employing the entire Deeplab V3+\cite{chen2018encoder} encoder as the embedding network following the standard approach, we demonstrate in Section~\ref{subsec:feature_lvl} that utilizing mid-level features of Deeplab V3+~\cite{chen2018encoder} results in improved transferability.

Then, we simply map the support and query features into two metric spaces and generate projected support and query features for each space with a fully-connected projection layer $f_{i}(\cdot)$, where $i\in\{{\rm cgs}, {\rm fgs}\}$ represents the coarse-grained space (blue area in Figure~\ref{fig:main}, segmenting salient human foreground) and the fine-grained space (green area in Figure~\ref{fig:main}, segmenting fine-grained human classes), respectively. In the following, we handle the parsing in two metric spaces concurrently through the momentum-updated prototypes and metric learning methods.

\vspace{-1em}
\subsection{Momentum-updated Prototypes}
\label{subsec:dp}

First, we employ the prototype learning~\cite{wang2019panet, dong2018few} approach that learns the representative prototypes for each class from the support features according to the ground truth support mask. However, there are very large appearance variances for the same fashion item class across the dataset. Therefore, for the same class, the static prototypes that each derived from a single support image are likely to have strike differences, which reduces the training stability. Thus, we generate the momentum-updated prototypes that are accumulated from the training time static prototypes to stabilize the training and learn robust features (which will be analyzed in Section~\ref{subsec:dp_exp}).

Specifically, in the coarse-grained metric space, we first generate static prototypes for a general ``human foreground'' and ``human background''. We denote the static prototypes for the ``human background'' and the ``human foreground'' as $\boldsymbol{p}_{{\rm cgs},0}$ and $\boldsymbol{p}_{{\rm cgs},1}$. The prototypes that are extracted from the projected features in the coarse-grained metric space can be formulated as:
\begin{equation}
\boldsymbol{p}_{{\rm cgs},c} = \frac{1}{{\left| {{\Lambda}_{c}^{s}} \right|}}\sum\limits_{x \in {\Lambda}_c^{s}} {f_{\rm cgs}({\boldsymbol{G}^{s}_{x}})},
\label{eq:pix2proto_cg}
\end{equation}
where $c\in \{0, 1\}$, ${\boldsymbol{G}^{s}_{x}}$ is the feature at pixel index $x$, ${\Lambda}_c^{s}$ is the index set for pixels in the $c$-th class in the support mask, and $\left| \cdot \right|$ measures the size of a set. Similarly, in the fine-grained metric space, we formulate the prototype for the $c$-th fine-grained class $\boldsymbol{p}_{{\rm fgs}, c}\in \mathbb{R}^D$ as:
\begin{equation}
\boldsymbol{p}_{{\rm fgs},c} = \frac{1}{{\left| {{\Lambda}_{c}^{s}} \right|}}\sum\limits_{x \in {\Lambda}_{c}^{s}} {{f_{\rm fgs}(\boldsymbol{G}^{s}_{x}}}),
\label{eq:pix2proto}
\end{equation}
where $c\in [1, |C_s|-1]$. Note that in the previous methods \cite{dong2018few, wang2019panet}, a ``background'' prototype is learned in the fine-grained metric space to represent non-foreground regions. Considering in the fine-grained metric space, we have no explicit knowledge about the novel classes and they are naturally labelled as the ``background'' class in the support mask, we do not explicitly calculate the static prototypes for the ``background'' class. Instead, we predict the background by learning a linear layer to exclude all the foreground classes that we will describe later in Section~\ref{subsec:agm} and Section~\ref{subsec:npm}. In the following sections, without loss of generality, we take the fine-grained
metric space as the example and omit the notations ${\rm cgs}$ and ${\rm fgs}$ in all $f(\cdot)$ and $\boldsymbol{p}$.

Next, instead of using a static prototype $\boldsymbol{p}_c$, we generate momentum-updated prototype $\boldsymbol{p}^d_c$ as the base class representations. Specifically, it is calculated by gradually smoothing $\boldsymbol{p}^d_c$ with the static prototype $\boldsymbol{p}_c$ in each training episode, $i.e.$,
\begin{equation}
\boldsymbol{p}^d_c \leftarrow \alpha \boldsymbol{p}^d_c + (1-\alpha)\boldsymbol{p}_c,
\label{eq:dynamicPrototype}
\end{equation}
where $\alpha\in [0, 1)$ is the momentum coefficient. In this way, the prototypes for the base classes are saved as network parameters and are available during meta-testing. Since the novel classes are not annotated during training, we use static prototypes for the novel classes during testing. For simplicity, unless specified, we denote the prototypes for both base and novel classes as $\boldsymbol{p}$ in the following sections.

\vspace{-1em}
\subsection{Dual Metric Learning}
\label{subsec:dml}

\begin{figure}[t]
  \centering
  \includegraphics[width=\linewidth]{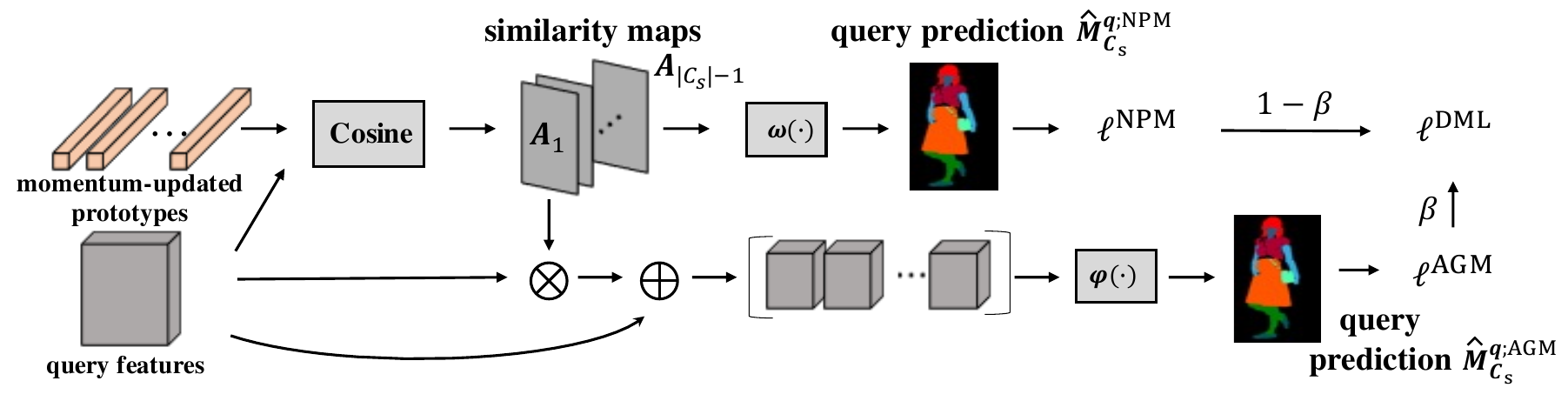}
  \vspace{-1.5em}
  \caption{Illustration of the dual metric learning (DML) scheme. Specifically, given the query features and the momentum-updated prototypes, we first calculate the cosine similarity (Cosine) and derive the similarity maps $\{\boldsymbol{A}_1, ..., \boldsymbol{A}_{|C_s|-1}\}$. In the early training phase, the network mainly focuses on optimizing an Attention Guidance Module (AGM) in the lower branch that predicts with the heavy $\varphi(\cdot)$. In the late training phase, the network shifts its focus to optimize a Nearest Prototype Module (NPM) in the upper branch where we directly predict from the similarity maps with the light-weight $\omega(\cdot)$. See Section~\ref{subsec:dml} for details.}\vspace{-1em}
  \label{fig:dml}
\end{figure}

Given the embedding query image features ${f_{\rm fgs}(\boldsymbol{G}^{s}})$ and the momentum-updated prototypes, the next step is to explore the query-support 
correlations and learn a good metric with DML, as depicted in Figure~\ref{fig:dml}.

In OS3, most SOTA methods seek to employ a set of trainable parameters to model the correlation and refine the query features with, \eg, pixel-level support distance attentions~\cite{zhang2018sg, gairola2020simpropnet}, pixel-level graph attentions~\cite{zhang2019pyramid, cao2020few}, appending pooled class prototypes~\cite{zhang2019canet}, and fusing prior masks produced by pre-trained high-level features~\cite{tian2020prior}. In a nutshell, the extra parameters are employed to learn the query-support correlations and bring pixels of the same class closer in the feature space. However, the novel classes are annotated as the ``background'' class in the OSHP task. Therefore, the extra parameters learn to push the novel class representations to be non-separable from the background representation and tend to overfit the base classes. In contrast to the mentioned methods, PANet~\cite{wang2019panet} comes up with a non-parametric structure that directly assigns the query pixels with the labels of the nearest support class prototype. Nevertheless, the simple non-parametric design cannot learn representative feature representations and discriminate the human classes.

In this paper, we show that integrating the two categories of approaches can obtain strong discriminative capability on the human classes and avoid encouraging the testing bias issue. Specifically, we propose to employ a DML scheme that includes a heavy AGM module (Section~\ref{subsec:agm}), a lightweight NPM module (Section~\ref{subsec:npm}), and a weight-shifting strategy (Section~\ref{subsec:weight_shift}) that seamlessly combines the two modules. The overview of our DML scheme is depicted in Figure~\ref{fig:dml}.

\subsubsection{Attention Guidance Module}
\label{subsec:agm}
In AGM, we learn to refine the attended pixel features in $f({\boldsymbol{G}^{q}})$ that correlates most with the prototypes and obtain more discriminative feature representations in the early training phase. In particular, we first derive the similarity maps for the $c$-th class by calculating cosine similarity between $f({\boldsymbol{G}^{q}})$ and the prototypes
\begin{equation}
\boldsymbol{A}_c= <{f({\boldsymbol{G}^{q}})}, \boldsymbol{p}_c>,
\label{eq:cosine}
\end{equation}
We then apply the Hadamard product between each similarity map and the query features to generate attended features for each class followed by a residual connection
\begin{equation}
\boldsymbol{R}_c = \boldsymbol{A}_{c} \otimes {f({\boldsymbol{G}^{q}})} + {f({\boldsymbol{G}^{q}})}. 
\label{eq:agm_res}
\end{equation}
Thus, we can get $|C_s| - 1$ residual features in total corresponding to the semantic classes, except for the ``background'' class. Then, we employ several separable convolutional layers akin to~\cite{chollet2017xception} denoted as $\varphi(\cdot)$ and $\varphi_{bg}(\cdot)$ to generate the probability scores $\boldsymbol{l}_c$, $i.e.$, 
\begin{equation}
\begin{small}
    \begin{aligned}
    \boldsymbol{l}_{c} &= \varphi(\boldsymbol{R}_c), \\
    \boldsymbol{l}_{0} &= \left(1/(|C_s| - 1\right) \times \sum_{c=1}^{|C_s-1|} \varphi_{bg}(\boldsymbol{R}_c).
    \end{aligned}
\label{eq:agm2}
\end{small}
\end{equation}
Note that we predict the probability scores that pixels do not belong to the $c$-th class with $\varphi_{bg}(\boldsymbol{R}_c)$ and explicitly model the probability for the ``background'' class by averaging the scores for the foreground classes. Finally, we generate the predicted probabilities $\hat{\boldsymbol{M}}_{c}^{q;{\rm AGM}}$ in AGM on the $c$-th class with a softmax layer
\begin{equation}
    \hat{\boldsymbol{M}}_{c}^{q;{\rm AGM}} = \frac{{\rm exp}(\boldsymbol{l}_c)}{\sum_{c=1}^{|C_s|-1}{\rm exp}(\boldsymbol{l}_c) + {\rm exp}(\boldsymbol{l}_{0})}.
\end{equation}
The AGM prediction is formulated by concatenating the predicted probabilities $\hat{\boldsymbol{M}}_{c}^{q;{\rm AGM}}$ for $c\in C_s$ and get $\hat{\boldsymbol{M}}^{q;{\rm AGM}}_{C_s}$. Finally, we supervise the AGM prediction with the ground truth query mask using cross-entropy loss $\ell^{{\rm AGM}}$.

\subsubsection{Nearest Prototype Module}
\label{subsec:npm}
$\varphi^{\rm AGM}$ and $\varphi^{\rm AGM}_{bg}$ employ a large number of parameters to model the query-support correlations and predict from the embedded features. However, the parameters can be easily overfitted to the base classes, leading to the testing bias issue described previously. Hence, we propose another lightweight NPM that directly infers the probability map for the $c$-th class from the similarity map $\boldsymbol{A}_c$ in the late training phase. The benefits of employing NPM are two folds. Firstly, it employs fewer parameters, which has high transferability to the novel classes. Secondly, a well-trained DML scheme only predicts from NPM. Therefore, it also boosts the inference speed by reducing computations.

In particular, we first derive the similarity for the ``background'' class by considering the areas that are not similar to any of the prototypes in the support set. Therefore, with the similarity map $\boldsymbol{A}_c$, we can explicitly calculate $1-\boldsymbol{A}_c$ as the dissimilarity map to the foreground class prototypes. Then, we fuse the averaged dissimilarity maps for all foreground classes and derive $\boldsymbol{A}_{0}$ for the ``background'' class, $i.e.$, 

\begin{equation}
\begin{small}
    \begin{aligned}
     \boldsymbol{A}_{0} = \big((1/(|C_s| - 1) \times \sum_{c=1}^{|C_s|-1} (1 - \boldsymbol{A}_c)\big).
    \end{aligned}
\label{eq:npm}
\end{small}
\end{equation}
In the next step, we learn two simple fully-connected projection layers $\omega(\cdot)$ and $\omega_{bg}(\cdot)$ taking the similarity maps $\boldsymbol{A}_c$ and $\boldsymbol{A}_{0}$ as input to infer the probabilities that which class the pixels belong to. And finally, we employ a softmax layer to get the NPM probability map $\hat{\boldsymbol{M}}_{c}^{q;{\rm NPM}}$ for the query image, \ie,
\begin{equation}
    \hat{\boldsymbol{M}}_{c}^{q;{\rm NPM}} = \frac{{\rm exp}\big({\omega}(\boldsymbol{A}_c)\big)}{\sum_{c=1}^{|C_s-1|}{\rm exp}\big({\omega}(\boldsymbol{A}_c)\big) + {\rm exp}\big({\omega_{bg}}(\boldsymbol{A}_{0})\big)}.
\end{equation}
Similar to the AGM prediction, the NPM prediction $\hat{\boldsymbol{M}}^{q;{\rm NPM}}_{C_s}$ is also formulated by concatenating the predicted probabilities for classes in $C_s$, which is supervised by the ground truth query mask with the cross-entropy loss $\ell^{{\rm NPM}}$.

\subsubsection{Weight-shifting Strategy}
\label{subsec:weight_shift}
In general, in the early training phase with AGM, we learn accurate similarity maps which are employed as class-level attention to enhance features' representational capability. In the late training phase with NPM, we directly predict from the similarity maps which are lightweight and have higher transferability to the novel classes. To model the phase changing from AGM to NPM, we propose seamlessly integrating the two modules by slowly shifting the network's focus via the weight-shifting strategy without any further fine-tuning. Note that in the coarse-grained metric space, we only employ AGM instead of DML since there are no novel classes that require strong transferability.

Formally, we can define the metric learning loss $\ell^{\rm DML}$ as:

\begin{equation}
\begin{small}
    \begin{aligned}
     \ell^{{\rm DML}} &= \beta\times\ell^{{\rm AGM}} + \left( 1 - \beta \right)\times\ell^{{\rm NPM}},
    \end{aligned}
\label{eq:weight}
\end{small}
\end{equation}
where $\beta$ is a balancing hyperparameter.
During training, we gradually shift the focus of the meta-learner by assigning loss weights that change along with the training epoch. We linearly decrease $\beta$ by $\beta = 1 - \rm{epoch / max\_epoch}$. After training, $\beta$ becomes $0$ and we solely infer $\hat{\boldsymbol{M}}_{C_s}^{q}$ from NPM.

\vspace{-1em}
\subsection{Prototype-level Contrastive Learning}
\label{subsec:contrastive}
According to~\cite{qi2018low}, the above metric learning approach actually can be seen as accumulating the contrastive InfoNCE/NCA loss~\cite{goldberger2004neighbourhood} between each pixel and the corresponding prototype. Unlike instance-level contrastive learning~\cite{he2020momentum, chen2020simple} with stable features, employing contrastive loss on individual pixels would introduce many noisy supervision signals, which hinder learning discriminative prototypes. To this end, we employ an auxiliary prototype-level contrastive loss which remedies the side effect of noisy pixels.

Formally, we extract the query class static prototype $\Tilde{\boldsymbol{p}}_{c}$ using the ground truth query mask during meta-training similar to Eq.~\eqref{eq:pix2proto}, $i.e.$,
\begin{equation}
\Tilde{\boldsymbol{p}}_{c} = \frac{1}{{\left| {{\Lambda}_{c}^{q}} \right|}}\sum\limits_{x \in {\Lambda}_{c}^ {q}} {{f({\boldsymbol{G}^{q}_{x}})}},
\label{eq:pix2proto2}
\end{equation}
where ${\Lambda}_c^{q}$ is the index set for pixels in the $c$-th class in the query mask and $\boldsymbol{G}^{q}_{x}$ is the feature at pixel index $x$. We next calculate the prototype-level contrastive loss with the extracted query static prototypes and the support static prototypes in Eq.~\eqref{eq:pix2proto}, $i.e.$,

\begin{equation}
\begin{small}
\ell^{{\rm NCA}}=\frac{1}{|C_s| - 1}\sum_{c=1}^{|C_s| - 1}(-{\rm log}\frac{{\rm exp}(<\Tilde{\boldsymbol{p}}_{c}, \boldsymbol{p}_c>/\tau)}{\sum_{k=1}^{|C_s-1|}{\rm exp}(<\Tilde{\boldsymbol{p}}_{c}, \boldsymbol{p}_{k}>/\tau)}),
\label{eq:nca_loss}
\end{small}
\end{equation}
where $\tau$ is the temperature coefficient. 
Finally, the overall objective function for the two spaces is defined as:
\begin{equation}
    \ell = \lambda^{{\rm NCA}}\ell^{{\rm NCA}} + \lambda^{{\rm AGM}}_{\rm cgs}\ell^{{\rm AGM}}_{\rm cgs} + \lambda^{{\rm DML}}_{\rm fgs}\ell^{{\rm DML}}_{\rm fgs},
\label{eq:objective}
\end{equation}
where $\lambda^{{\rm NCA}}$, $\lambda^{{\rm AGM}}_{\rm cgs}$, and $\lambda^{{\rm DML}}_{\rm fgs}$ are balancing hyperparameters.

\vspace{-1em}
\subsection{Discussion}

To recall, POPNet~\cite{he2021progressive} builds a progressive refinement pipeline to tackle the OSHP problem that gradually infuses the coarse-grained structural knowledge into the fine-grained features in three stages. In this way, the model progressively focuses on the target classes at the finer granularity with rich parent semantics. There are two main technical differences between EOP-Net and POPNet~\cite{he2021progressive}. First, the progressive architecture proposed by POPNet~\cite{he2021progressive} adopts inconsistent training strategies for different stages, where only the fine-grained stages employ metric learning. Therefore, the parameters for each stage are trained separately in sequence. On the contrary, EOP-Net employs an end-to-end architecture to parse humans into human classes of different granularities in two metric spaces that share the same embedding network. In this way, the parameters in the metric spaces can be optimized simultaneously while enjoying fewer parameters, less computational complexity, less overall training time, and higher inference speed. The end-to-end design makes our EOP-Net more computationally efficient and easier to train. Detailed analysis of the model complexity can be found in Section~\ref{subsec:complex}.
We empirically find that
the end-to-end architecture design also boosts the performance significantly thanks to the more discriminative embedding network. The performance gain for the architecture design is further discussed in Section~\ref{subsec:progress_end2end}. Second, the prototype-level contrastive loss between the query and support prototypes is proposed in EOP-Net, which remedies the noisy pixel issue and improves the quality of the momentum-updated prototypes. We evaluate the effectiveness of the prototype-level contrastive loss in Section~\ref{subsec:abl_contrast}.

\vspace{-1em}
\section{Datasets}
In this section, we present the details on how to tailor the existing large-scale human parsing datasets into new one-shot datasets for the OSHP setting.

\vspace{-1em}
\subsection{ATR-OS} 
\label{subsec:ATR}
ATR dataset~\cite{liang2015, liang2015human} is a large-scale single human parsing benchmark including 16,000 training images annotated with 17 foreground classes. Images in the ATR dataset are mostly fashion photographs, including models and a variety of fashion items, which are closely related to the applications of OSHP, such as fashion clothing parsing \cite{yamaguchi2012parsing}. Whereas the pose, size, and position of models in the ATR dataset are less diverse than the other human parsing datasets.

Specifically, we first split the ATR images into the support sets and the query sets for meta-training and meta-testing, respectively. We adopt the original train/val data split in ATR dataset to split the meta-training/meta-testing in ATR-OS. For meta-training, we first form $\mathcal{Q}_{train}$ by including the first 7,500 images of the ATR training set and form $\mathcal{S}_{train}$ with the remaining images. For evaluation, we draw 1,650 query-support pairs to form the meta-testing set from the ATR validation set and ensure that each class is evaluated at least 150 times on different query-support pairs.

To ease the difficulty of training OSHP on the ATR dataset, we merge the symmetric classes and rare classes in ATR, $e.g.$ ``left leg'' and ``right leg'' are merged as ``legs'' and ``sunglasses'' is merged into ``background''. The remaining 12 classes including ``background'' are denoted as $C_{human}$. The semantic classes in the human parsing task can be categorized into human body classes and fashion item classes. Since human body classes like ``legs'' and ``arms'' have the same definition in most of the applications, we put all human body classes into the base classes $C_{base}$ and divide the remaining fashion item classes into two novel class sets representing two main body areas, respectively, $i.e.$, the lower-body area: $C_{Fold\ 1}$ = [``dress'', ``skirt'', ``pants''] and the upper-body area: $C_{Fold\ 2}$ = [``hat'', ``upper-clothes'', ``bag'']. Before training, we select one fold as $C_{novel}$ and merge the classes in the other fold into $C_{base}$. Note that the fold formulation is different from~\cite{he2021progressive} and the new formulation is more suitable for real-world applications that require replacing and re-defining the lower-body/upper-body fashion classes.

Since only classes in $C_{base}$ are exposed during meta-training and we have no extra information for the novel classes, we merge the $C_{novel}$ segments into the ``background'' class in $S_{train}$ and $Q_{train}$. During meta-testing, all classes in $C_{human}$ will be evaluated, including classes from both $C_{base}$ and $C_{novel}$.

\vspace{-1em}
\subsection{LIP-OS} LIP dataset~\cite{liang2018look} is another large-scale single human parsing dataset with more than 30,000 training images. Unlike ATR, images from LIP are collected from real-world scenarios and differ greatly in pose, view, appearance, and resolution, making it more challenging than the ATR dataset.

Similar to the tailoring process in ATR-OS, we split LIP training samples into 14,000 images for $\mathcal{Q}_{train}$ and 16,462 images for $\mathcal{S}_{train}$. Then we form 1,800 meta-testing pairs to evaluate each class at least 150 times. The 20 classes in LIP are also merged into 13 classes. We select the two folds in LIP-OS with the same logic in ATR-OS. The two folds are $C_{Fold\ 1}$ = [dress, skirt, pants] constituting the lower-body area and $C_{Fold\ 2}$ = [upper-clothes, coat, jumpsuit] constituting the upper-body area. Before training, we select classes from one fold as $C_{novel}$ and label the novel class segments as the ``background'' class in the meta-training set.

\vspace{-1em}
\subsection{CIHP-OS}
We have also tailored the CIHP dataset into the one-shot setting. The CIHP dataset is a multi-human parsing dataset with 3.4 averaged instances per image. Like the LIP dataset, CIHP images also contain high human appearance variability and complexity, making this dataset more challenging than the ATR dataset.

Similarly, in the tailored CIHP-OS dataset, the 28,280 training images in CIHP are divided into 14,000 images for $\mathcal{Q}_{train}$ and 14,280 images for $\mathcal{S}_{train}$. The meta-testing list is built with 1,800 query-support pairs, enabling each class to be evaluated at least 150 times. The fashion class are selected in two folds: $C_{Fold\ 1}$ = [dress, skirt, pants] and the upper-body area: $C_{Fold\ 2}$ = [upper-clothes, hat, coat]. Before training, one fold is selected as $C_{novel}$ while the other classes serve as $C_{base}$, and the class in $C_{novel}$ are labelled as the ``background'' class.
%%%%%%%%%%%%%%%%ATR%%%%%%%%%%%%%%%

\begin{table*}[!t]
\centering
\renewcommand{\arraystretch}{1.2}
\caption{Comparisons on the ATR-OS dataset. ``Ave'' denotes the averaged results from two folds. ``Bi-mIoU'' denotes binary mIoU.}\vspace{-1em}
\label{tab:atr}
\resizebox{\textwidth}{!}{%
\begin{tabular}{cccccccc|ccccccc}
\hline
\multicolumn{1}{c|}{\multirow{3}{*}{Method}} & \multicolumn{7}{c|}{$K$-way OSHP} & \multicolumn{7}{c}{$1$-way OSHP} \\ \cline{2-15} 
\multicolumn{1}{c|}{} & \multicolumn{3}{c|}{$C_{novel}$ mIoU (\%)} & \multicolumn{3}{c|}{$C_{human}$ mIoU (\%)} & \multirow{2}{*}{Overall Acc. (\%)} & \multicolumn{3}{c|}{$C_{novel}$ mIoU (\%)} & \multicolumn{3}{c|}{$C_{human}$ mIoU (\%)} & \multirow{2}{*}{Bi-mIoU (\%)} \\
\multicolumn{1}{c|}{} & Fold 1 & Fold 2 & \multicolumn{1}{c|}{Ave} & Fold 1 & Fold 2 & \multicolumn{1}{c|}{Ave} &  & Fold 1 & Fold 2 & \multicolumn{1}{c|}{Ave} & Fold 1 & Fold 2 & \multicolumn{1}{c|}{Ave} &  \\ \hline
Fine-tuning~\cite{caelles2017one} & - & - & - & - & - & - & - & 0.0 & 0.0 & 0.0 & 13.3 & 12.0 & 12.7 & 48.7 \\
PerSAM~\cite{zhang2023personalize} & - & - & - & - & - & - & - & 9.4 & 3.3 & 4.2 & 6.9 & 6.9 & 6.9 & 14.6 \\
PANet~\cite{wang2019panet} & 0.4 & 1.7 & 1.1 & 6.0 & 4.3 & 5.2 & 48.5 & 9.2 & 8.2 & 8.7 & 16.6 & 16.0 & 16.3 & 49.4 \\
PPNet~\cite{zhang2020part} & - & - & - & - & - & - & - & 37.8 & 31.1 & 34.5 & 58.5 & 59.9 & 59.2 & 76.0 \\
SG-One~\cite{zhang2018sg} & 0.3 & 0.7 & 0.5 & 54.8 & 54.1 & 54.5 & 91.8 & 11.5 & 10.6 & 11.1 & 56.9 & 61.0 & 59.0 & 73.6 \\
AMP~\cite{siam2019amp} & 9.4 & 8.3 & 8.9 & 20.5 & 19.6 & 20.1 & 69.1 & 7.6 & 5.6 & 6.6 & 12.8 & 9.3 & 11.1 & 49.1 \\
PFENet~\cite{tian2020prior} & - & - & - & - & - & - & - & 18.8 & 11.4 & 15.1 & 43.5 & 43.2 & 43.4 & 67.3 \\
POPNet~\cite{he2021progressive} & 29.0 & 29.3 & 29.2 & 55.3 & 56.8 & 56.1 & 91.9 & 40.3 & 26.7 & 31.9 & 58.4 & 66.9 & 62.3 & 69.8 \\
EOP-Net & \textbf{31.1} & \textbf{34.6} & \textbf{32.9} & \textbf{61.9} & \textbf{63.3} & \textbf{62.6} & \textbf{93.5} & \textbf{53.0} & \textbf{41.4} & \textbf{47.2} & \textbf{68.2} & \textbf{69.5} & \textbf{68.9} & \textbf{83.2} \\ \hline
\end{tabular}%
}\vspace{-1em}
\end{table*}

%%%%%%%%%%%%%%%%LIP%%%%%%%%%%%%%%%

\begin{table*}[!t]
\renewcommand{\arraystretch}{1.2}
\centering
\caption{Comparisons on the LIP-OS dataset. ``Ave'' denotes the averaged results from two folds. ``Bi-mIoU'' denotes binary mIoU.}\vspace{-1em}
\label{tab:lip}
\resizebox{\textwidth}{!}{%
\begin{tabular}{cccccccc|ccccccc}
\hline
\multicolumn{1}{c|}{\multirow{3}{*}{Method}} & \multicolumn{7}{c|}{$K$-way OSHP} & \multicolumn{7}{c}{$1$-way OSHP} \\ \cline{2-15} 
\multicolumn{1}{c|}{} & \multicolumn{3}{c|}{$C_{novel}$ mIoU (\%)} & \multicolumn{3}{c|}{$C_{human}$ mIoU (\%)} & \multirow{2}{*}{Overall Acc. (\%)} & \multicolumn{3}{c|}{$C_{novel}$ mIoU (\%)} & \multicolumn{3}{c|}{$C_{human}$ mIoU (\%)} & \multirow{2}{*}{Bi-mIoU (\%)} \\
\multicolumn{1}{c|}{} & Fold 1 & Fold 2 & \multicolumn{1}{c|}{Ave} & Fold 1 & Fold 2 & \multicolumn{1}{c|}{Ave} &  & Fold 1 & Fold 2 & \multicolumn{1}{c|}{Ave} & Fold 1 & Fold 2 & \multicolumn{1}{c|}{Ave} &  \\ \hline
Fine-tuning~\cite{caelles2017one} & - & - & - & - & - & - & - & 0.0 & 0.0 & 0.0 & 13.7 & 10.1 & 11.9 & 49.1 \\
PerSAM~\cite{zhang2023personalize} & - & - & - & - & - & - & - & 13.7 & 19.2 & 16.5 & 10.0 & 10.0 & 10.0 & 9.8 \\
PANet~\cite{wang2019panet} & 0.0 & 0.0 & 0.0 & 5.0 & 5.0 & 5.0 & 48.7 & 9.6 & 8.6 & 9.1 & 13.2 & 12.1 & 12.7 & 43.5 \\
PPNet~\cite{zhang2020part} & - & - & - & - & - & - & - & 36.4 & 23.7 & 30.1 & 43.1 & 42.3 & 42.7 & 62.9 \\
SG-One~\cite{zhang2018sg} & 2.8 & 2.3 & 2.6 & 29.5 & 31.9 & 30.7 & 75.6 & 18.0 & 15.7 & 16.9 & 43.0 & 41.0 & 42.0 & 61.9 \\
AMP~\cite{siam2019amp} & 11.0 & 11.8 & 11.4 & 13.7 & 13.7 & 13.7 & 30.7 & 4.1 & 7.1 & 5.6 & 10.0 & 11.7 & 10.9 & 46.3 \\
PFENet~\cite{tian2020prior} & - & - & - & - & - & - & - & 0.0 & 16.6 & 8.3 & 6.9 & 24.5 & 15.7 & 50.1 \\
POPNet~\cite{he2021progressive} & 22.1 & 29.7 & 25.9 & 31.9 & 31.8 & 31.9 & 72.9 & 36.9 & \textbf{48.9} & 42.9 & 52.3 & 56.3 & 54.3 & 66.1 \\
EOP-Net & \textbf{25.7} & \textbf{30.4} & \textbf{28.1} & \textbf{43.0} & \textbf{45.7} & \textbf{44.4} & \textbf{80.9} & \textbf{42.0} & 46.2 & \textbf{44.1} & \textbf{57.0} & \textbf{58.0} & \textbf{57.5} & \textbf{75.8} \\ \hline
\end{tabular}%
}\vspace{-1em}
\end{table*}

\begin{table*}[!t]
\renewcommand{\arraystretch}{1.2}
\centering
\caption{Comparisons on the CIHP-OS dataset. ``Ave'' denotes the averaged results from two folds. ``Bi-mIoU'' denotes binary mIoU.}\vspace{-1em}
\label{tab:cihp}
\resizebox{\textwidth}{!}{%
\begin{tabular}{cccccccc|ccccccc}
\hline
\multicolumn{1}{c|}{\multirow{3}{*}{Method}} & \multicolumn{7}{c|}{$K$-way OSHP} & \multicolumn{7}{c}{$1$-way OSHP} \\ \cline{2-15} 
\multicolumn{1}{c|}{} & \multicolumn{3}{c|}{$C_{novel}$ mIoU (\%)} & \multicolumn{3}{c|}{$C_{human}$ mIoU (\%)} & \multirow{2}{*}{Overall Acc. (\%)} & \multicolumn{3}{c|}{$C_{novel}$ mIoU (\%)} & \multicolumn{3}{c|}{$C_{human}$ mIoU (\%)} & \multirow{2}{*}{Bi-mIoU (\%)} \\
\multicolumn{1}{c|}{} & Fold 1 & Fold 2 & \multicolumn{1}{c|}{Ave} & Fold 1 & Fold 2 & \multicolumn{1}{c|}{Ave} &  & Fold 1 & Fold 2 & \multicolumn{1}{c|}{Ave} & Fold 1 & Fold 2 & \multicolumn{1}{c|}{Ave} &  \\ \hline
Fine-tuning~\cite{caelles2017one} & - & - & - & - & - & - & - & 0.0 & 0.0 & 0.0 & 7.2 & 7.1 & 9.3 & 46.3 \\
PerSAM~\cite{zhang2023personalize} & - & - & - & - & - & - & - & 7.8 & 10.3 & 9.1 & 7.1 & 7.1 & 7.1 & 10.4 \\
PANet~\cite{wang2019panet} & 1.1 & 0.7 & 0.9 & 4.0 & 5.4 & 4.7 & 39.9 & 11.1 & 9.2 & 10.2 & 15.5 & 13.4 & 14.5 & 48.3 \\
PPNet~\cite{zhang2020part} & - & - & - & - & - & - & - & 16.1 & 27.3 & 21.7 & 43.4 & 44.7 & 44.1 & 67.2 \\
SG-One~\cite{zhang2018sg} & 0.8 & 0.3 & 0.5 & 42.9 & 37.1 & 40 & 81.8 & 7.9 & 10.0 & 9.0 & 47.0 & 45.4 & 46.2 & 65.4 \\
AMP~\cite{siam2019amp} & 8.9 & 9.7 & 9.3 & 20.4 & 22.5 & 21.5 & 56.3 & 5.5 & 5.3 & 5.4 & 9.2 & 10.3 & 9.8 & 46.6 \\
PFENet~\cite{tian2020prior} & - & - & - & - & - & - & - & 10.2 & 3.1 & 6.7 & 29.0 & 22.5 & 25.8 & 58.7 \\
POPNet~\cite{he2021progressive} & 18.5 & 16.9 & 17.7 & 49.0 & 37.2 & 43.1 & 79.5 & \textbf{31.0} & \textbf{40.3} & \textbf{35.7} & \textbf{56.5} & \textbf{58.4} & \textbf{57.7} & 70.9 \\
EOP-Net & \textbf{20.5} & \textbf{25.1} & \textbf{22.8} & \textbf{49.1} & \textbf{45.5} & \textbf{47.3} & \textbf{84.0} & 25.4 & 36.4 & 30.9 & 53.8 & 55.4 & 54.6 & \textbf{74.4} \\ \hline
\end{tabular}%
}\vspace{-1em}
\end{table*}

\vspace{-1em}
\section{Experiments}
\subsection{Implementation Details}
In this paper, we train and evaluate our models on a single NVIDIA Tesla V100 GPU with 16 GB memory. The input images are augmented by first resizing to $512 \times 512$, then applying a random scale from 0.5 to 2, a random crop, and a random flip. We train our models using Stochastic Gradient Descent (SGD) optimizer for 50 epochs with the poly learning rate policy. The initial learning rate is set to 0.001 with a batch size of 2. When generating momentum-updated prototypes, the momentum coefficient $\alpha$ in Eq.~\eqref{eq:dynamicPrototype} is set to 0.999 by grid search. However, static prototypes are utilized when calculating similarity maps in the first $3$ epochs before the feature representations are stable enough to reduce the variance of the static prototypes. The temperature hyperparameter $\tau$ for the contrastive loss in Eq.~\eqref{eq:nca_loss} is set to 0.1. In the final objective function, $\lambda^{\rm{NCA}}$, $\lambda^{\rm{AGM}}_{\rm cgs}$, and $\lambda^{\rm{DML}}_{\rm fgs}$ are set to 1.0.

\vspace{-1em}
\subsection{Metrics} 
We evaluate parsing in two OSHP settings. The main setting is $k$-way OSHP that parses $k$ human classes in one episode. For $k$-way OSHP, we use mean Intersection over Union (mIoU) as the main metric for evaluating the parsing performance on the classes in $C_{novel}$ and $C_{human}$. We additionally use the overall accuracy averaged from all pixels to measure the overall human parsing performance. In the other OSHP setting named $1$-way OSHP, we evaluate the scenario when only one human class is required to be parsed. For $1$-way OSHP, in addition to mIoU, we also compute the average Binary-IoU~\cite{wang2019panet} to evaluate the general capability of discriminating foreground from the background.

\vspace{-1em}
\subsection{Contenders} 
To comprehensively evaluate the effectiveness of the proposed method, we implement the representative OS3 methods into the OSHP settings with the same augmentation techniques as our EOP-Net described as follows.

\noindent{\textbf{Fine-tuning~\cite{caelles2017one}}} is a classic method that fine-tunes the model on the meta-test dataset. To implement Fine-tuning~\cite{caelles2017one} for OSHP, we first pre-train the model on $\mathcal{Q}_{train}$ and $\mathcal{S}_{train}$ then fine-tune it on the $\mathcal{S}_{test}$ for a few iterations. Specifically, we use the same backbone as EOP-Net and only fine-tune the last two convolutional layers and the classification layer. 

\noindent{\textbf{SG-One~\cite{zhang2018sg}}} in Section~\ref{subsec:rel_os3}. We follow the exact settings of SG-One~\cite{zhang2018sg} except for replacing the backbone with the same one as EOP-Net for better performance. Since the original SG-One~\cite{zhang2018sg} does not support $k$-way OSHP, we follow a similar prediction procedure as defined by Eq.~\eqref{eq:agm2} in our AGM to generate the prediction for the ``background'' class except not using residual features.

\noindent{\textbf{PANet~\cite{wang2019panet}}} in Section~\ref{subsec:dml}. We implement PANet~\cite{wang2019panet} as a baseline of non-parametric metric learning with a prototype alignment loss. Because PANet requires $k$ support images and $k$ support masks to parse one query image in the $k$-way OSHP, we pair each query image with $k$ support images that each contains a unique class (\ie, with a binary support mask) as described in \cite{wang2019panet} during meta-training. In meta-testing, we use the same support image for each query-support pair but generate $k$ masks by selecting one class at a time and masking the other classes as the ``background'' class. Due to the high memory consumption for PANet~\cite{wang2019panet}, we use its original ResNet-50 backbone~\cite{he2016deep} to save GPU memory. 

\noindent{\textbf{AMP~\cite{siam2019amp} }} utilizes masked proxies with multi-resolution weight imprinting to generate segmentation results. We carefully tune AMP~\cite{siam2019amp} model to a suitable learning rate as described in~\cite{siam2019amp}. However, as the multi-resolution imprinting described in~\cite{siam2019amp} cannot be fully applied to the Xception backbone, we keep its original backbone.

\noindent{\textbf{PPNet~\cite{zhang2020part}}} employs part-aware prototypes that are decomposed from the static prototypes. However, since one scene contains multiple human classes and we lack the global semantic class information as described in PPNet~\cite{zhang2020part}, we have to remove the semantic branch in~\cite{zhang2020part}. Also, since PPNet~\cite{zhang2020part} employs a more complex scheme than PANet~\cite{wang2019panet}, it has even higher memory consumption. Therefore, we use the original ResNet-50 backbone~\cite{he2016deep} and only carry out PPNet~\cite{zhang2020part} in $1$-way OSHP.

\noindent{\textbf{PFENet~\cite{tian2020prior}}} in Section~\ref{subsec:dml}. When comparing EOP-Net to PFENet~\cite{tian2020prior}, we only tune the network to a proper learning rate and use the same hyperparameters and structures that are described in~\cite{tian2020prior}. We train PFENet~\cite{tian2020prior} with the ResNet-50~\cite{he2016deep} backbone, which has the highest performance according to~\cite{tian2020prior}.

\noindent{\textbf{POPNet~\cite{he2021progressive}.}} We compare our method with our previous method POPNet~\cite{he2021progressive} as well. For a fair comparison, we replace the original embedding network in POPNet~\cite{he2021progressive} with the embedding network described in Section~\ref{subsec:end2end}. Since the dataset settings in ATR-OS have been changed and are no longer in a cluster-disjoint manner, we discard the second stage in POPNet~\cite{he2021progressive} and directly infuse the first-stage knowledge into the third stage.

\noindent\textbf{PerSAM~\cite{zhang2023personalize}}. Built upon the recent powerful Segment Anything Model (SAM)~\cite{kirillov2023segment}, PerSAM complements SAM to allow segmenting a unique visual concept that is defined by a single reference image and mask. To do so, PerSAM takes positive and negative location priors within the reference and iteratively aggregates the features within the foreground target regions. Since PerSAM needs a careful design to be extended to k-way OSHP, we only compare it with EOP-Net under the 1-way OSHP setting. We employ the pretrained SAM model with a ViT-H encoder.

\begin{figure*}[t]
  \centering
  \includegraphics[width=\linewidth]{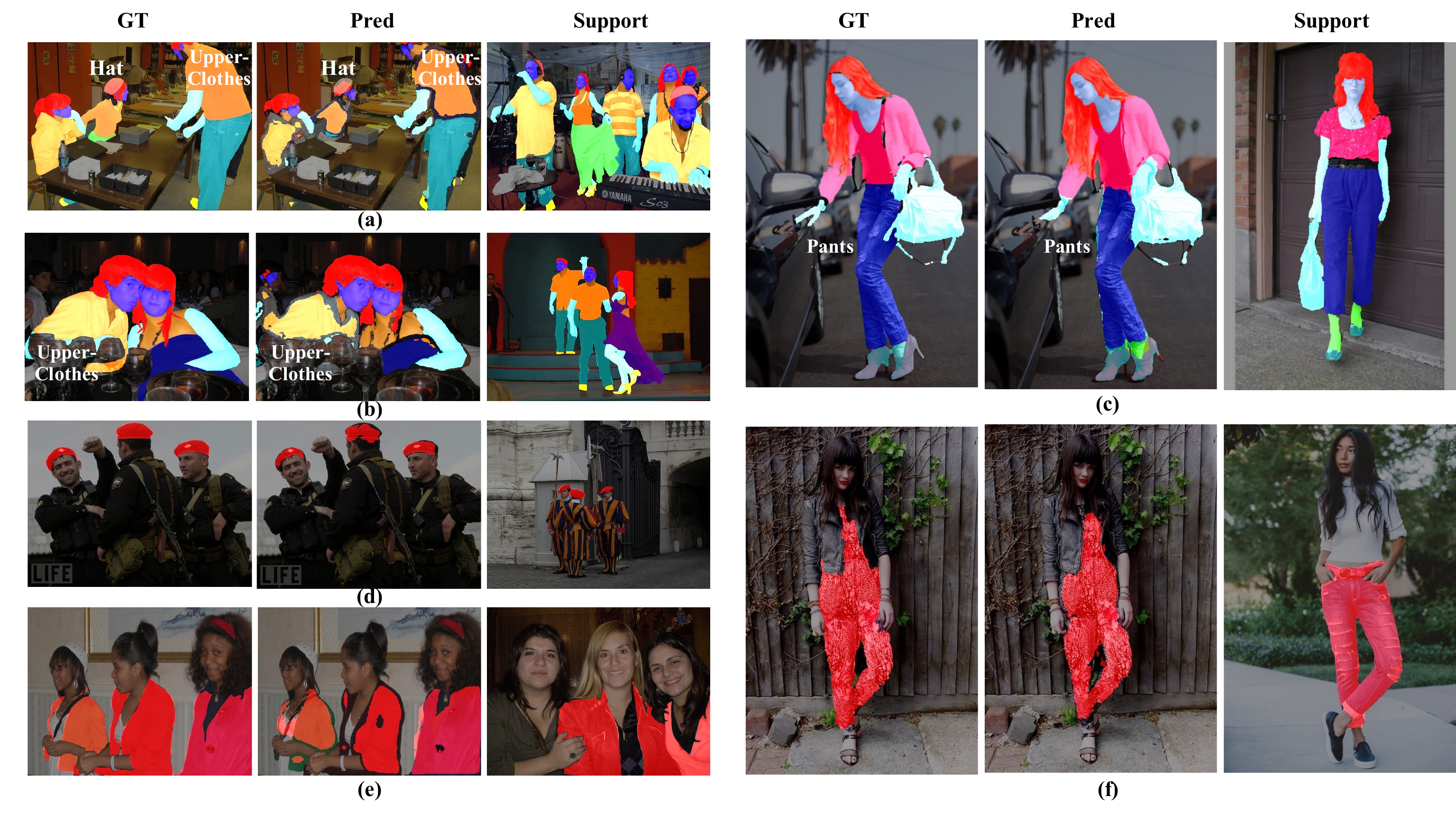}
  \vspace{-2.5em}
  \caption{Visual results of EOP-Net on the CIHP-OS and ATR-OS meta-testing set. (a)-(c) $k$-way OSHP. (d)-(f) $1$-way OSHP. The novel classes in $k$-way OSHP have white text labels and all parsed classes in $1$-way OSHP are the novel classes. ``Upper-clothes'' is in both yellow/orange due to color's addition effect between the background images and the semantic maps.}\vspace{-1em}
  \label{fig:kw1w}
\end{figure*}

\begin{figure*}[!t]
    \centering
\includegraphics[width=\linewidth]{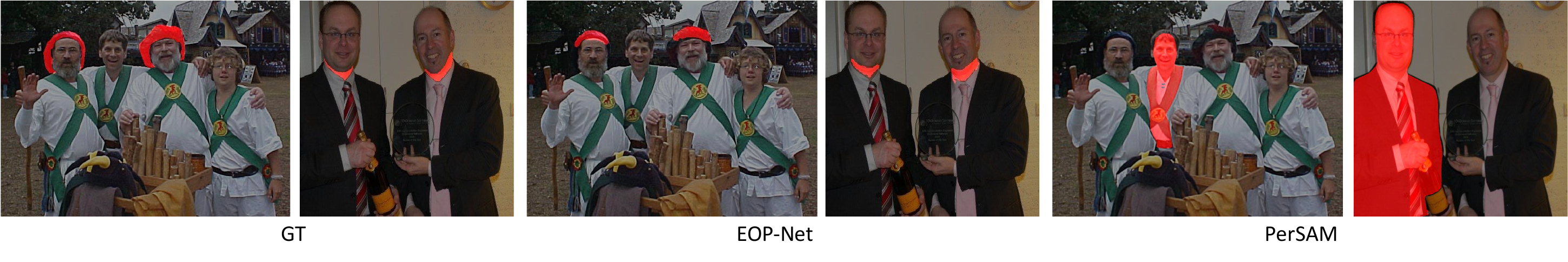}
\vspace{-2.5em}
    \caption{Qualitative comparison between EOP-Net and PerSAM~\cite{zhang2023personalize} on CIHP-OS Fold 1.}\vspace{-1em}
    \label{fig:persam}
\end{figure*}

\vspace{-0.5em}
\subsection{Quantitive Results}
\label{subsec:expMain}

We compare our model EOP-Net with POPNet~\cite{he2021progressive} and six customized OS3 baselines in Table~\ref{tab:atr}, Table~\ref{tab:lip}, and Table~\ref{tab:cihp} on ATR-OS, LIP-OS, and CIHP-OS datasets respectively (we compare the mIoU averaged from the two folds).

We observe that our proposed EOP-Net outperforms the contenders by large margins, including the preliminary version POPNet~\cite{he2021progressive}. For example, EOP-Net achieves a gain of 5.1\% and 4.2\% on the $C_{novel}$ mIoU and $C_{human}$ mIoU for $k$-way OSHP from POPNet~\cite{he2021progressive} on the CIHP-OS dataset (Table~\ref{tab:cihp}). The performance gain comes from applying the end-to-end human parsing framework that shares semantic information across different granularities and employing the auxiliary prototype-level contrastive that separates the similar human classes in the feature space. We also empirically find that the SOTA method PerSAM largely falls behind EOP-Net on all datasets. Furthermore, through visualizations in Figure~\ref{fig:persam} (Figure~5 of the revised manuscript), we observe that although PerSAM has strong segmentation capabilities, it still struggles with the challenging OSHP task and produces trivial solutions covering the entire human body instead of accurately segmenting fine-grained human parts. We conjecture that due to OSHP's challenges of small sizes and similar parts, PerSAM fails to build good human class representations. In contrast, our EOP-Net enhances these representations with specific designs of the end-to-end human parsing framework and prototype-level contrastive learning and thus obtains better results. In the future, we can incorporate PerSAM with effective designs in our EOP-Net and parameter-efficient fine-tuning to better exploit the SAM foundation models and generate robust OSHP predictions. 

However, on the CIHP-OS dataset, POPNet~\cite{he2021progressive} gets better results for $1$-way OSHP. The reason is that CIHP-OS contains more challenging poses and occlusions with multiple persons in a scene, which makes learning the human foreground in the coarse-grained metric space challenging. POPNet~\cite{he2021progressive} hence has more advantages for employing more parameters in different stages. In contrast, for $k$-way OSHP, since the semantic information for $k$ classes is provided, EOP-Net can better utilize the extra semantic information to refine the human foreground with the end-to-end human parsing framework and achieve better performance.

Note that there is a performance gap when comparing the $1$-way OSHP scores to $k$-way counterparts for most methods since parsing $k$ classes together would introduce more interventions and confusion among the different human classes. Also, the gap between $1$-way OSHP scores and $k$-way is higher on $C_{novel}$ mIoU compared to $C_{human}$ mIoU. The reason is that the model would be less confident in the novel classes when the base classes are involved during testing. However, providing the semantic information for $k$ classes helps our model learn the underlying semantic relations across different human classes and can improve the overall human parsing performance.

\begin{figure*}[t]
  \centering
  \includegraphics[width=\linewidth]{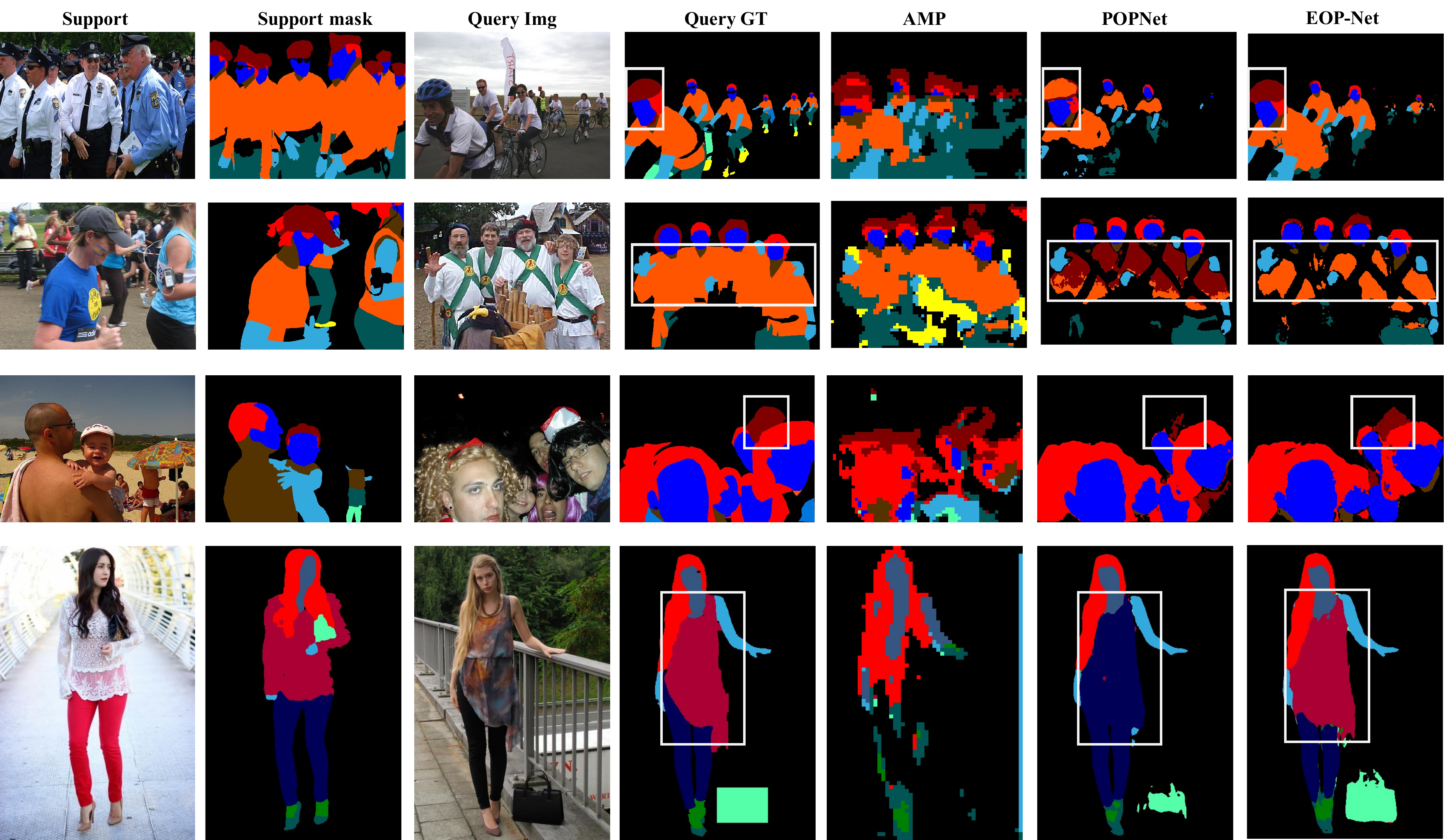}\vspace{-1em}
  \caption{Visual results of different methods for $k$-way OSHP on the CIHP-OS (top two rows) and ATR-OS (bottom two rows) meta-testing set. Our methods POPNet~\cite{he2021progressive} and EOPNet generate more accurate predictions than the best-performed baseline method AMP~\cite{siam2019amp}. The differences are highlighted in white boxes.}\vspace{-1em}
  \label{fig:main_vis}
\end{figure*}

\vspace{-0.5em}
\subsection{Qualitative Results}
\subsubsection{General Visual Inspection}
To better understand the OSHP task and the effectiveness of EOP-Net, we further show the qualitative results generated by EOP-Net on both $k$-way and $1$-way OSHP settings in Figure~\ref{fig:kw1w}. Generally, we observe that EOP-Net can flexibly generate satisfying parsing masks for classes defined by the support example, including the novel classes that are not annotated in the training data.
For example, for the $k$-way OSHP on CIHP-OS in Figure~\ref{fig:kw1w} (a) and (b), although the scenes contain crowded humans in various challenging poses, EOP-Net is still able to generate high-quality parsing results in terms of both the small-size base classes, \eg, ``face'' and ``arms'', and the novel classes, \eg, ``hat'' and ``upper-clothes'' in (a). For the $1$-way OSHP on CIHP-OS, since the model is only required to learn to parse one target class which is a relatively easier setting, the quality of the parsing results is better than that for the $k$-way OSHP. On ATR-OS in Figures (c) and (f), since the human foreground is more salient, the parsing results are more precise.

\subsubsection{Visual Comparison with Baselines}
\label{subsec:visual_comp}
With the same support image and support mask, we compare our EOP-Net with the baseline methods for $k$-way OSHP. As shown in Figure~\ref{fig:main_vis}, 
although AMP~\cite{siam2019amp} yields the highest scores on $C_{novel}$ mIoU among the baselines, it still struggles to locate and discriminate the novel classes, indicating that the OSHP task is indeed challenging for the OS3 methods. On the contrary, POPNet~\cite{he2021progressive} and our EOP-Net can successfully discriminate the fine-grained human classes, including the novel ones. When comparing EOP-Net with POPNet~\cite{he2021progressive}, we can observe that EOP-Net has higher accuracy for discriminating base and novel classes, as indicated by the white boxes. We speculate that the end-to-end human parsing framework and the prototype-level contrastive loss learn more discriminative feature representations and separate the similar parts in the metric space, respectively.

We also notice some failure patterns in Figure~\ref{fig:main_vis}. For instance, in the first and second rows, both POPNet~\cite{he2021progressive} and our EOP-Net fail to parse the humans far away from the camera because these areas have less encoded local semantic information. 
We leave fixing the problem as the future work.

\begin{figure*}[t]
  \centering
  \includegraphics[width=\linewidth]{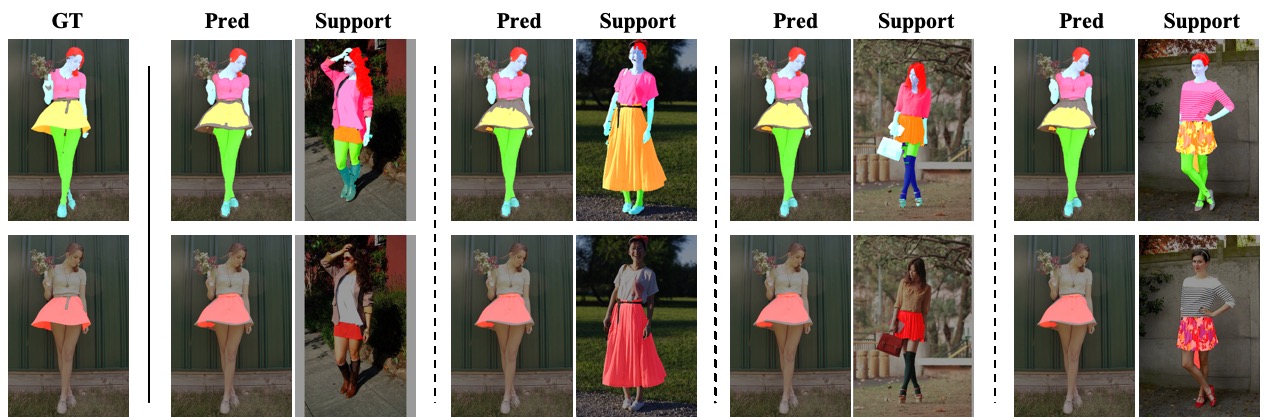}\vspace{-1em}
  \caption{Visual results for parsing the same query image with different support pairs on ATR-OS meta-testing set. The novel class ``skirt'' is labeled in yellow/orange in $k$-way OSHP and red color in $1$-way OSHP.}\vspace{-1.5em}
  \label{fig:stable}
\end{figure*}
\subsubsection{Parsing Stability}
To evaluate the stability of EOP-Net for OSHP, we map the same query image with different support pairs sampled from the ATR-OS meta-testing set in Figure~\ref{fig:stable}. Although there is a noticeable variation in the appearance of the support images, our method can still generate relatively stable parsing results on the novel class ``skirt''. These results indicate that the momentum-updated prototypes for the base classes facilitate learning robust features, which is transferable to the novel classes despite the appearance variations in the support images.

\begin{table}
\caption{Ablation study results on the ATR-OS dataset Fold 1. ``WS'', ``CG \& FG'', ``CL'' are short for the weight shifting strategy, parsing both coarse-grained and fine-grained human classes with the end-to-end human parsing framework, and adding the auxiliary prototype-level contrastive loss.}\vspace{-1em}
\resizebox{\linewidth}{!}{
\centering
\small
\begin{tabular}{@{}c|cc@{}}
Methods   & $C_{novel}$ mIoU (\%) & $C_{human}$ mIoU (\%) \\ \hline
AGM  & 0.4  & 51.6 \\
NPM  & 8.2 & 16.5 \\
DML w/o WS & 13.1 & 33.3 \\
DML & 21.6 & 58.6 \\       
DML + CG \& FG  & 30.2 & 61.6 \\
DML + CG \& FG + CL & 31.1 & 61.9
\end{tabular}%
}\vspace{-1em}
\label{tab:ablation}
\end{table}

\vspace{-1em}
\subsection{Ablation Studies}

\begin{figure}[t]
  \centering
  \includegraphics[width=0.8\linewidth]{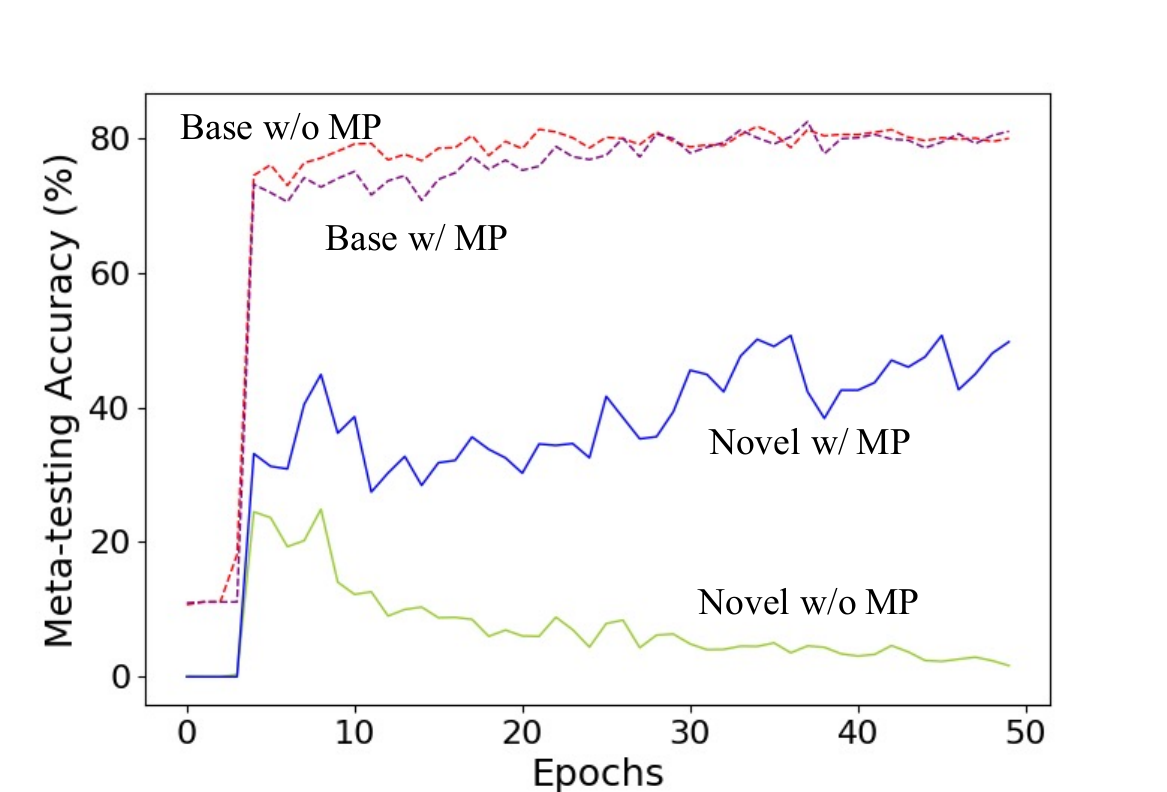}\vspace{-1em}
  \caption{Training on ATR-OS dataset Fold 1. ``Base w/ MP'' and ``Base w/o MP''  denotes the base classes with and without momentum-updated prototypes, respectively. ``Novel w/ MP'' and ``Novel w/o MP''  denotes the novel classes with and without momentum-updated prototypes, respectively.}\vspace{-1em}
  \label{fig:abl_dp}
\end{figure}

We investigate the effectiveness of key components of our model on the ATR-OS dataset Fold 1 in Table~\ref{tab:ablation} and Figure~\ref{fig:abl_dp}. 

\subsubsection{DML Scheme}
Firstly, we can observe from the first row in Table~\ref{tab:ablation} that solely applying AGM ($\beta=1$ during training in Eq.~\eqref{eq:weight}) has an acceptable $C_{human}$ mIoU but very low $C_{novel}$ mIoU. It indicates that the model overfits the base classes. Then, when solely applying NPM ($\beta=0$ during training in Eq.~\eqref{eq:weight}) in the second row, we can observe that although the $C_{novel}$ mIoU is higher, the model has an overall low performance for both $C_{novel}$ mIoU and $C_{human}$ mIoU. The reason is that the model cannot discriminate the fine-grained similar human classes with a lightweight NPM structure. The experimental results support the intuition for developing DML in Section~\ref{subsec:dml}. By employing both modules in the third row without the weight-shifting strategy ($\beta=0.5$ during training in Eq.~\eqref{eq:weight}), the $C_{novel}$ mIoU
can be improved to 13.1\%, suggesting that improving the metric quality by modeling the query-support correlations with AGM and maintaining a lightweight module with NPM with high transferability are both important for EOP-Net. When further adding the weight shifting strategy in the fourth row, our model can significantly improve $C_{novel}$ mIoU and $C_{human}$ mIoU by large margins, $i.e.$, achieving 21.6\% and 58.6\%, respectively. It's indicated that integrating AGM with NPM via the weight-sharing strategy can further boost performance for our EOP-Net. 

\vspace{-0.5em}
\subsubsection{End-to-end Human Paring Framework}\label{subsec:progress_end2end}
The above ablation experiments are only conducted on the fine-grained metric space. We then formulate parsing with the proposed
end-to-end human parsing framework to parse both coarse-grained and fine-grained human classes. We can observe that $C_{novel}$ mIoU and $C_{human}$ mIoU are improved by 8.6\% and 3.0\%, implying that mutually refining feature representations of different granularities can derive a stronger embedding network.

We also compare our end-to-end human parsing framework with the progressive human parsing framework in POPNet~\cite{he2021progressive} in terms of the parsing performance. The comparison on ATR-OS Fold 1 is shown as follows:
\vspace{-0.5em}
\begin{center}
\resizebox{0.8\linewidth}{!}{
\centering
\vspace{-.2em}
\tablestyle{1.1pt}{1.1}
\begin{tabular}{c|cc}
Parsing framework & $C_{novel}$ mIoU (\%) & $C_{human}$ mIoU (\%) \\
\hline
Progressive & 29.0 & 55.3 \\
End-to-end & 30.2 & 61.1 \\
\end{tabular}
\vspace{-1em}
}
\end{center}
We notice that replacing the progressive parsing framework with the end-to-end parsing framework can boost the $C_{novel}$ mIoU and $C_{human}$ mIoU by 1.2\% and 5.8\%, respectively, demonstrating
the superiority of the end-to-end  
solution
for having a higher ability to locate and discriminate the small-size human classes compared to the progressive framework.

\subsubsection{Prototype-level Contrastive Loss}\label{subsec:abl_contrast}
After applying the prototype-level contrastive loss on the prototypes, we can observe that EOP-Net finally reaches 31.1\% $C_{novel}$ mIoU and 61.9\% $C_{human}$ mIoU. It is noteworthy that the mIoU for $C_{base}$ reaches 71.2\%, which is close to the fully supervised human parsing methods. Together with the visual inspection results in Figure~\ref{fig:main_vis}, we speculate that applying the auxiliary prototype-level contrastive loss is beneficial for separating the base classes in the fine-grained space. We also surprisingly notice that such capability is also transferable to the novel classes.

\subsubsection{Momentum-updated Prototypes}
\label{subsec:dp_exp}
We investigate the training curve with and without the momentum-updated prototypes in Figure~\ref{fig:abl_dp}. We compare our method with and without momentum-updated prototypes in terms of the mean $C_{base}$ accuracy and the mean $C_{novel}$ accuracy for the ATR-OS dataset. Interestingly, although we only have the momentum-updated prototypes for the base classes, the mean $C_{base}$ accuracy becomes similar when training with/without momentum-updated prototypes after training for 30 epochs. On the contrary, mean $C_{novel}$ accuracy shows an improving trend when equipped with momentum-updated prototypes and vice versa. To further investigate this phenomenon, we report the mIoU with different values of the momentum coefficient $\alpha$ (Eq.~(3)) on ATR-OS Fold 1 in Table~\ref{tab:alpha}. We find that generally higher momentum coefficients (more robust base class presentations) lead to better $C_{novel}$ mIoU. We speculate that robust base-class representations brought by momentum-updated prototypes stabilize the training process to learn more representative features for the novel classes. Thus, we set $\alpha=0.999$ as the default setting.

\begin{table}[!t]
\centering
\small
\caption{Effect of the momentum coefficient $\alpha$ on ATR-OS Fold 1.}
\vspace{-1em}
\label{tab:alpha}
\resizebox{0.3\textwidth}{!}{%
\begin{tabular}{@{}c|ccccc@{}}
$\alpha$ & 0.8 & 0.9 & 0.99 & 0.999 & 0.9999 \\ \hline
$C_{novel}$ mIoU & 25.7 & 28.3 & 30.7 & 31.1 & 31.0 \\
$C_{human}$ mIoU & 60.6 & 61.2 & 61.2 & 61.9 & 61.7 \\
\end{tabular}%
}
\vspace{-1em}
\end{table}

\subsubsection{Depth of Embedding Network} \label{subsec:feature_lvl}

In this paper, we employ a customized Deeplab V3+~\cite{chen2018encoder} encoder as the embedding network. The original Deeplab V3+ encoder has a total of 74 convolutional layers. Here, we investigate the effect of its depth. We report the results of EOP-Net under three common depth variants of 62, 69, and 74 layers in Table~\ref{tab:depth}. We observe that the $62$-layer embedding network achieves the highest $C_{novel}$ mIoU, suggesting that mid-level features have higher generalization capabilities than high-level features. Thus, we employ the $62$-layer embedding network as the default setting.

\begin{table}
\caption{Effect of the depth of the embedding network on ATR-OS Fold 1.}\vspace{-1em}
\centering
\resizebox{0.9\linewidth}{!}{
\small
\begin{tabular}{@{}c|cc@{}}
Embedding network  & $C_{novel}$ mIoU (\%) & $C_{human}$ mIoU (\%) \\ \hline
$62$-layer & 31.1 & 61.9 \\
$69$-layer & 30.5 & 61.7 \\
$74$-layer & 30.4 & 61.8 \\
\end{tabular}%
}\vspace{-1em}
\label{tab:depth}
\end{table}

\begin{table}[!t]
\centering
\small
\caption{Model complexity and running time of different methods.}\vspace{-1em}
\label{tab:complexity}
\resizebox{0.4\textwidth}{!}{%
\begin{tabular}{@{}c|ccc@{}}
Method & MACs (G) & Parameters (M) & FPS \\ \hline
PANet~\cite{wang2019panet} & 612.6 & 14.7 & 2.3 \\
PPNet~\cite{zhang2020part} & 560.0 & 24.7 & 0.3 \\
POPNet~\cite{he2021progressive} & 264.4 & 85.8 & 7.8 \\
EOP-Net & 129.3 & 42.5 & 12.0 \\
\end{tabular}%
}\vspace{-1em}
\end{table}

\vspace{-1em}
\subsection{Model Complexity and Running Time}\label{subsec:complex}
We compare the computational complexity, model size, and running time of our models with representative $k$-way OS3 models in Table~\ref{tab:complexity}. Images are at a resolution of 512$\times$512, and the support class number $k$ in each scene is 5. We measure the computational cost by multiply-and-accumulates (MACs), the model size by the number of parameters, and the inference speed by frames per second (FPS) with 
a batch size of 1. The FPS results are averaged over the first 30 images of ATR-OS. The average number of classes is 9.3. We observe that although the parameters for PANet~\cite{wang2019panet} and PPNet~\cite{zhang2020part} are less than our models, they have higher computational complexity and latency since they require loading $k$ different support images in one meta-training/meta-testing episode. In contrast, POPNet~\cite{he2021progressive} and EOP-Net can fully utilize $k$ classes annotated in one support image and hence 
reduce computation. Compared to POPNet~\cite{he2021progressive}, EOP-Net further reduces the computational cost and the parameters by half. The FPS is also improved to 12.0. The improved efficiency comes from sharing the embedding network for parsing human in different granularities with the end-to-end human parsing framework instead of employing the progressive framework as in~\cite{he2021progressive}.

\section{Conclusion}
In this work, we have introduced a new challenging but promising new task, $i.e.$, one-shot human parsing (OSHP) that requires parsing human images into an open set of classes defined by a single reference image, which can largely alleviate the annotation effort and benefit many fashion application scenarios.
To this end, we have devised a strong baseline, $i.e.$, End-to-end One-shot human Parsing Network (EOP-Net). EOP-Net can effectively learn a strong embedding network, develop representative features with high transferability, and separate the human classes in the metric space, thereby handling the small sizes, testing bias, and similar parts issues. Moreover, we have also established the first benchmark by tailoring three large-scale human parsing datasets to the OSHP settings, which can serve as a testbed for training and validating OSHP models. Extensive experimental results on the three datasets have demonstrated the effectiveness of EOP-Net and its superiority over the state-of-the-art one-shot semantic segmentation methods in terms of both the generalization ability on the novel classes and the overall parsing ability of the entire human body.

To address the failure cases in Section~\ref{subsec:visual_comp}, we plan to continue our research in the following directions. The first is to improve the local semantic reasoning ability by exploring the natural human class relationships through graph convolutional networks~\cite{wu2020comprehensive, wang2020hierarchical} or vision Transformers~\cite{deit, pan2021scalable, xu2021vitae}. The second is to combine one-shot human parsing with other human-centric analysis tasks to improve the perception of human body semantics, \eg, human pose estimation \cite{zhang2020towards}. It is also possible to improve the one-shot parsing performance on small human instances and body parts in crowded scenes by exploiting instance-aware bounding boxes~\cite{zhao2020fine, zhao2018understanding, li2018multi}. Moreover, clustering~\cite{ge2020self} and self-training can be helpful for explicitly modeling the relationship of novel classes.

\ifCLASSOPTIONcaptionsoff
  \newpage
\fi

\bibliographystyle{IEEEtran}
\bibliography{reference}

\vspace{-3em}
\begin{IEEEbiography}
[{\includegraphics[width=1in,height=1.25in,clip,keepaspectratio]{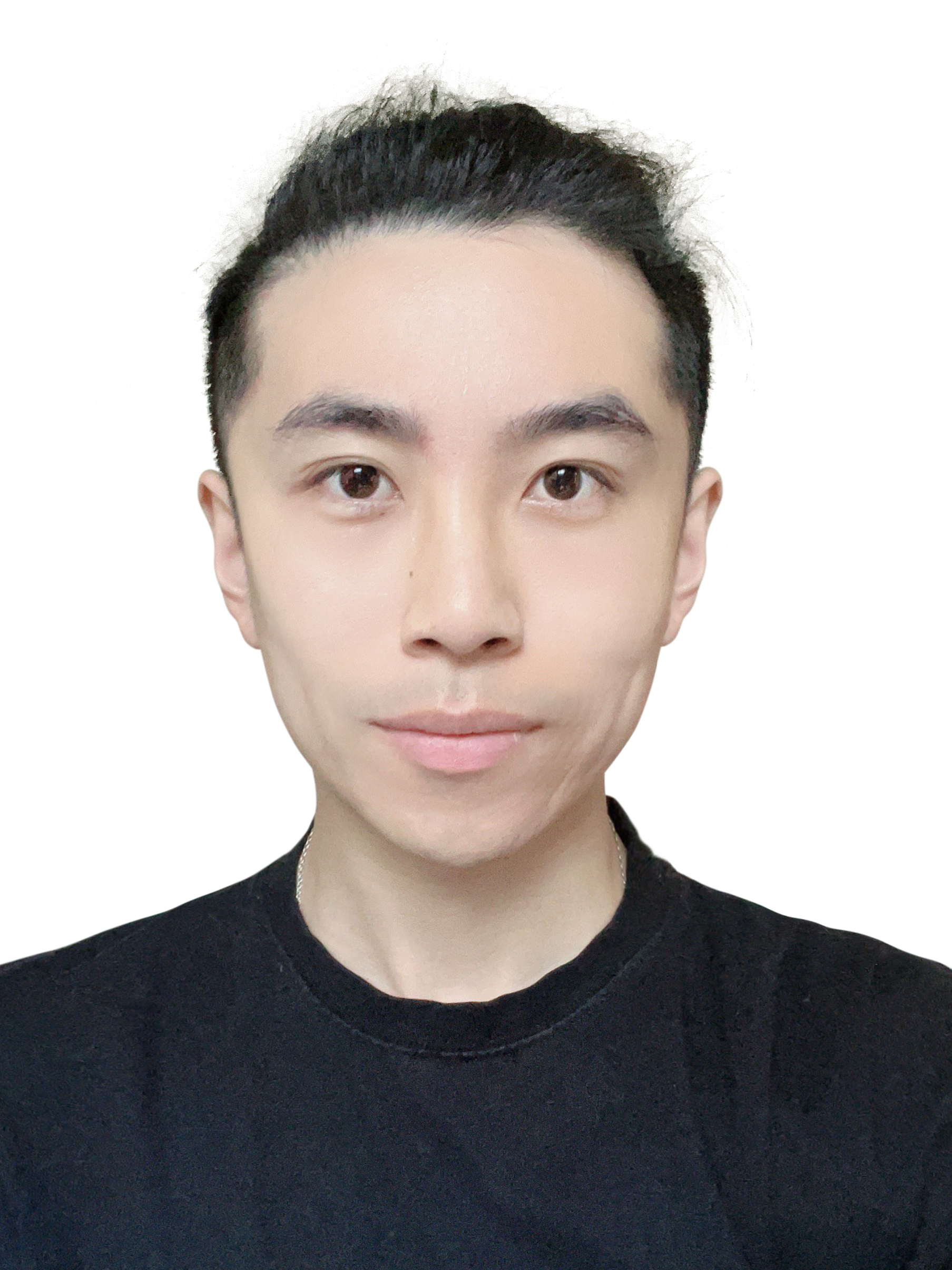}}]{Haoyu He}
is a Ph.D. student in the Faculty of Information Technology, Monash University Clayton Campus, Australia. He received his BCS and Mphil Degrees in 2019 and 2021, both from the University of Sydney, Australia.
His research interests include computer vision, efficient deployment of large models, model compression, and segmentation tasks.
\end{IEEEbiography}

\vspace{-1.5em}
\begin{IEEEbiography}[{\includegraphics[width=1in,height=1.25in,clip,keepaspectratio]{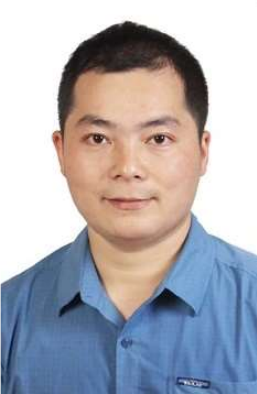}}]{Jing Zhang} (Senior Member, IEEE) is currently a Research Fellow at the School of Computer Science, The University of Sydney. He has published more than 60 papers in prestigious conferences and journals, such as CVPR, ICCV, ECCV, NeurlPS, ICLR, IEEE TPAMI, and IJCV. His research interests include computer vision and deep learning. He is also a Senior Program Committee Member of the AAAI Conference on Artificial Intelligence and the International Joint Conference on Artificial Intelligence. He serves as a regular reviewer for many prestigious journals and conferences.
\end{IEEEbiography}

\vspace{-1.5em}
\begin{IEEEbiography}[{\includegraphics[width=1in,height=1.25in,clip,keepaspectratio]{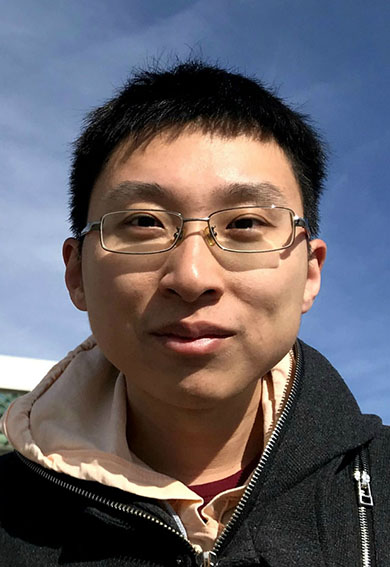}}]{Bohan Zhuang} is now a tenure-track assistant professor and doctoral supervisor at the Faculty of Information Technology, Monash University, Australia. He primarily focuses on efficient machine learning research, with a particular emphasis on model quantization and pruning, as well as designing lightweight neural architectures. He has published over 40 papers in top-tier international conferences (e.g., CVPR, NeurIPS) and journals (e.g., IEEE Transactions on Pattern Analysis and Machine Intelligence) in computer vision and machine learning venues. Part of the outcomes have been highly cited and translated into many edge-native AI tools by industry. He has served as the senior committee member of several renowned conferences including ICML, NeurIPS, ICLR, CVPR and ICCV.
\end{IEEEbiography}

\vspace{-1.5em}
\begin{IEEEbiography}[{\includegraphics[width=1in,height=1.25in,clip,keepaspectratio]{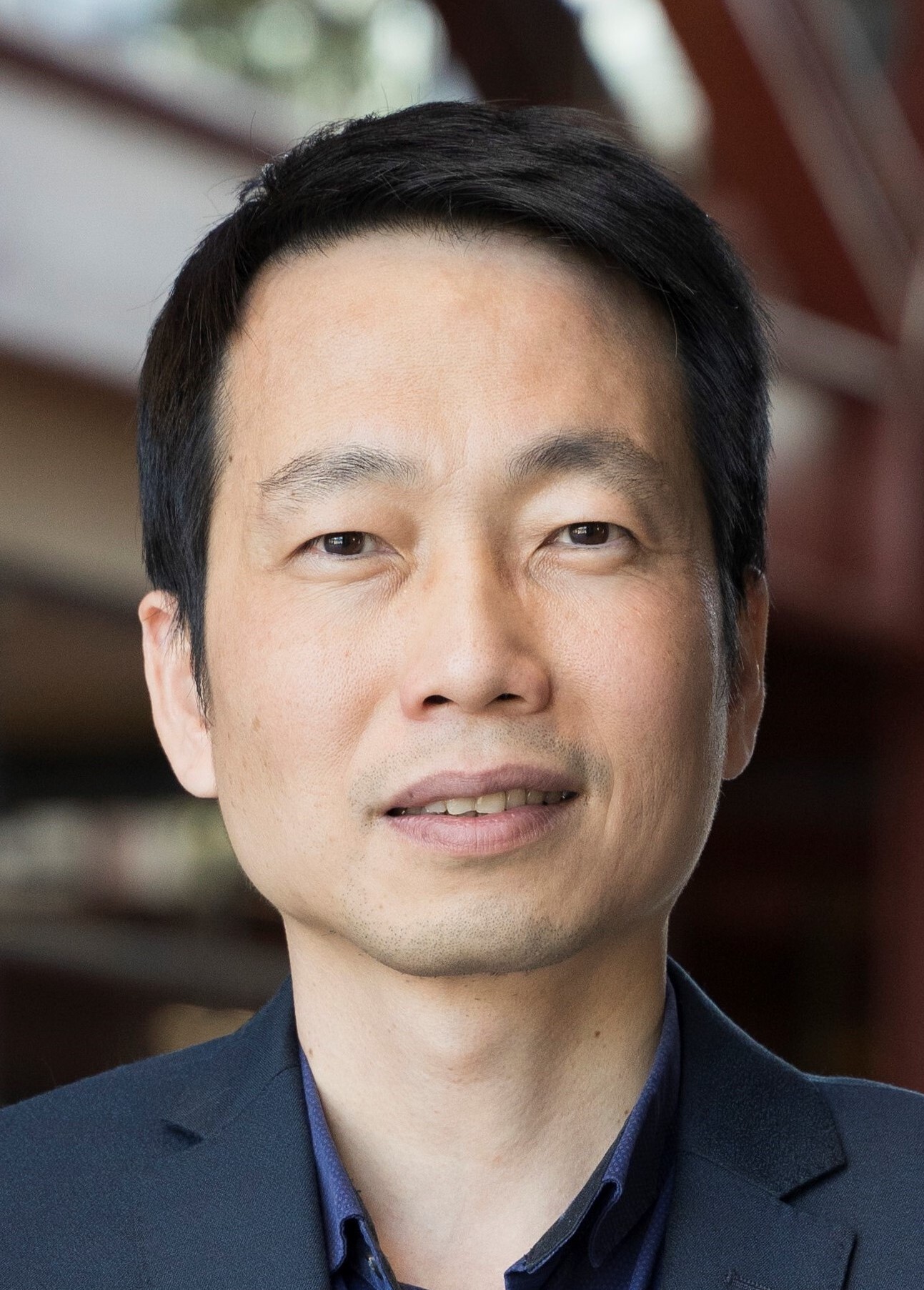}}]{Jianfei Cai}(S'98-M'02-SM'07-F’21) received his PhD degree from the University of Missouri-Columbia. He is currently a Professor and serves as the Head of the Data Science \& AI Department at Faculty of IT, Monash University, Australia. Before that, he had served as Head of Visual and Interactive Computing Division and Head of Computer Communications Division in Nanyang Technological University (NTU). His major research interests include computer vision, multimedia and visual computing. He is a co-recipient of paper awards in ACCV, ICCM, IEEE ICIP and MMSP. He serves or has served as an Associate Editor for TPAMI, IJCV, IEEE T-IP, T-MM, and T-CSVT as well as serving as Area Chair for CVPR, ICCV, ECCV, IJCAI, ACM Multimedia, ICME and ICIP. He was the Chair of IEEE CAS VSPC-TC during 2016-2018. He is the leading TPC Chair for IEEE ICME 2012 and the leading general chair for ACM Multimedia 2024. He is a Fellow of IEEE.
\end{IEEEbiography}

\vspace{-1.5em}
\begin{IEEEbiography}
[{\includegraphics[width=1in,height=1.25in,clip,keepaspectratio]{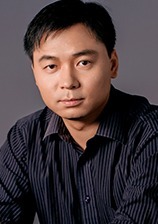}}]{Dacheng Tao} (Fellow, IEEE) is currently a Professor of Computer Science, Peter Nicol Russell Chair and an Australian Laureate Fellow in the Sydney AI Center and the School of Computer Science in the Faculty of Engineering at The University of Sydney. He mainly applies statistics and mathematics to artificial intelligence and data science, and his research is detailed in one monograph and over 200 publications in prestigious journals and proceedings at leading conferences. He received the 2015 and 2020 Australian Eureka Prize, the 2018 IEEE ICDM Research Contributions Award, and the 2021 IEEE Computer Society McCluskey Technical Achievement Award. He is a Fellow of the Australian Academy of Science, AAAS, ACM and IEEE. \end{IEEEbiography}

\end{document}